\documentclass{article}
\usepackage{iclr2017_conference,times}
\usepackage{hyperref}
\usepackage{url}
\usepackage[]{natbib}
\usepackage[]{algorithm2e}
\usepackage[toc,page]{appendix}
\usepackage{graphicx}
\usepackage{amsmath}
\usepackage{amssymb}
\usepackage{hyperref}
\usepackage{pgfplots}

\DeclareMathOperator*{\maxA}{\max}

\newcommand{\dd}{{\mathrm{d}}}
\newcommand{\pd}[2]{\frac{\partial #1}{\partial #2}}

\newcommand{\od}[2]{\frac{\dd #1}{\dd #2}}

\newcommand{\pp}[1]{\left( #1 \right)}

\newcommand{\mc}{\mathcal}
\newcommand{\argmin}{\operatornamewithlimits{argmin}}
\newcommand{\argmax}{\operatornamewithlimits{argmax}}

\newcommand{\norm}[1]{\left|\left| #1 \right|\right|}
\newcommand{\expect}[2]{\mathbb{E}_{#1}\left[ #2 \right]}

\newcommand{\jcom}[1]{}
\newcommand{\bcom}[1]{}

\newcommand{\lcom}[1]{}
\newcommand{\dcom}[1]{}

\title{Unrolled Generative Adversarial Networks}

\author{Luke Metz\thanks{Work done as a member of the Google Brain Residency program (\href{http://g.co/brainresidency}{g.co/brainresidency})}\\
Google Brain\\
\texttt{lmetz@google.com}
\And
Ben Poole\thanks{Work completed as part of a Google Brain internship}\\
Stanford University \\
\texttt{poole@cs.stanford.edu}
\And
David Pfau \\
Google DeepMind \\
\texttt{pfau@google.com} 
\And
Jascha Sohl-Dickstein\\
Google Brain\\
\texttt{jaschasd@google.com}
}

\iclrfinalcopy 
\begin{document}

\maketitle

\begin{abstract}
We introduce a method to stabilize Generative Adversarial Networks (GANs) by defining the generator objective with respect to an unrolled optimization of the discriminator. 
This allows training to be adjusted 
between using the optimal discriminator in the generator's objective, which is ideal but infeasible in practice, and using the current value of the discriminator, which is often unstable and leads to poor solutions. 
We show how this technique solves the common problem of mode collapse, stabilizes training of GANs with complex recurrent generators, and increases diversity and coverage of the data distribution by the generator.
\end{abstract}

\section{Introduction}

The use of deep neural networks as generative models for complex data has made great advances in recent years. 
This success has been achieved through a surprising diversity of training losses and model architectures, including denoising autoencoders \citep{Pascal10}, variational autoencoders \citep{Kingma13,Rezende2014,Gregor15,Kulkarni15,Burda15,Kingma16}, generative stochastic networks \citep{Alain15}, diffusion probabilistic models \citep{Jascha15}, autoregressive models \citep{Theis2015c,Oord2016a,Oord2016b}, real non-volume preserving transformations \citep{Dinh14,Dinh16}, Helmholtz machines \citep{Dayan95,bornschein2015bidirectional}, and Generative Adversarial Networks (GANs) \citep{Goodfellow14}.

\subsection{Generative Adversarial Networks}
\label{sec intro gan}
While most deep generative models are trained by maximizing log likelihood or a lower bound on log likelihood, GANs take a radically different approach that does not require inference or explicit calculation of the data likelihood. Instead, two models are used to solve a minimax game: a generator which samples data, and a discriminator which classifies the data as real or generated. In theory these models are capable of modeling an arbitrarily complex probability distribution. When using the optimal discriminator for a given class of generators, the original GAN proposed by Goodfellow et al. minimizes the Jensen-Shannon divergence between the data distribution and the generator, and extensions generalize this to a wider class of divergences \citep{Nowozin16,Sonderby16,poole2016improved}. 

The ability to train extremely flexible generating functions, without explicitly computing likelihoods or performing inference, and while targeting more mode-seeking divergences as made GANs extremely successful in image generation \citep{Odena2016b,Salimans16,Radford15}, and image super resolution \citep{Ledig2016}.
The flexibility of the GAN framework has also enabled a number of successful extensions of the technique, for instance for structured prediction \citep{reed2016learning,reed2016generative, Odena2016b}, training energy based models \citep{Zhao2016}, and combining the GAN loss with a mutual information loss \citep{Chen16}.

In practice, however, GANs suffer from many issues, particularly during training. 
One common failure mode involves the generator collapsing to produce only a single sample or a small family of very similar samples. 
Another involves the generator and discriminator oscillating during training, rather than converging to a fixed point.
In addition, if one agent becomes much more powerful than the other, the learning signal to the other agent becomes useless, and the system does not learn.
To train GANs many tricks must be employed, such as careful selection of architectures \citep{Radford15}, minibatch discrimination \citep{Salimans16}, and noise injection \citep{Salimans16,Sonderby16}. 
Even with these tricks the set of hyperparameters for which training is successful is generally very small in practice.

Once converged, the generative models produced by the GAN training procedure normally do not cover the whole distribution \citep{Dumoulin16,Che16}, even when targeting a mode-covering divergence such as KL. 
Additionally, because it is intractable to compute the GAN training loss, and because approximate measures of performance such as Parzen window estimates suffer from major flaws \citep{Theis2016a}, evaluation of GAN performance is challenging. 
Currently, human judgement of sample quality is one of the leading metrics for evaluating GANs. In practice this metric does not take into account mode dropping if the number of modes is greater than the number of samples one is visualizing. In fact, the mode dropping problem generally helps visual sample quality as the model can choose to focus on only the most common modes. These common modes correspond, by definition, to more typical samples. Additionally, the generative model is able to allocate more expressive power to the modes it does cover than it would if it attempted to cover all modes.

\subsection{Differentiating Through Optimization}

Many optimization schemes, including SGD, RMSProp \citep{Tieleman2012}, and Adam \citep{Kingma14}, consist of a sequence of differentiable updates to parameters. Gradients can be backpropagated through unrolled optimization updates in a similar fashion to backpropagation through a recurrent neural network. The parameters output by the optimizer can thus be included, in a differentiable way, in another objective \citep{Maclaurin15}. This idea was first suggested for minimax problems in \citep{Pearlmutter08}, while \citep{Zhang10} provided a theoretical analysis and experimental results on differentiating through a single step of gradient ascent for simple matrix games. Differentiating through unrolled optimization was first scaled to deep networks in \citep{Maclaurin15}, where it was used for hyperparameter optimization. More recently, \citep{Belanger15,Han16,andrychowicz2016learning} backpropagate through optimization procedures in contexts unrelated to GANs or minimax games.

In this work we address the challenges of unstable optimization and mode collapse in GANs by unrolling optimization of the discriminator objective during training.

\section{Method}

\subsection{Generative Adversarial Networks}

The GAN learning problem is to find the optimal parameters $\theta_G^*$ for a generator function $G\pp{ z; \theta_G}$ in a minimax objective, 
\begin{align} \label{eq:base_obj}
\theta_G^* &= \argmin_{\theta_G} \maxA_{\theta_D} f\pp{\theta_G, \theta_D} \\
&= \argmin_{\theta_G} f\pp{\theta_G, \theta_D^*\pp{\theta_G}} \label{eq:gan loss}\\
\theta_D^*\pp{\theta_G} &= \argmax_{\theta_D}  f\pp{\theta_G, \theta_D}
,
\end{align}
where $f$ is commonly chosen to be
\begin {align}
f\pp{\theta_G, \theta_D} &= 
    \expect{x\sim p_{data}}{\mathrm{log}\pp{D\pp{x; \theta_D}}} +
    \expect{z \sim \mathcal{N}(0,I)}{\mathrm{log}\pp{1 - D\pp{G\pp{z; \theta_G}; \theta_D}}}
    .
    \label{eq: f minimax}
\end{align}
Here $x \in \mc X$ is the data variable, $z \in \mc Z$ is the latent variable, $p_{data}$ is the data distribution, the discriminator $D\left(\cdot; \theta_D \right): \mc X \to [0,1]$ outputs the estimated probability that a sample $x$ comes from the data distribution, $\theta_D$ and $\theta_G$ are the discriminator and generator parameters, and the generator function $G\pp{ \cdot; \theta_G}: \mc Z \to \mc X$ transforms a sample in the latent space into a sample in the data space. 

For the minimax loss in Eq. \ref{eq: f minimax}, the optimal discriminator $D^*\left(x \right)$ is a known smooth function of the generator probability $p_G\left(x\right)$ \citep{Goodfellow14},
\begin {align}
D^*\left(x \right) &= 
    \frac{
        p_{data}\pp{x}
        }{
        p_{data}\pp{x} + p_G\left(x\right)
        }
    .
\label{eq: true D*}
\end{align}
When the generator loss in Eq. \ref{eq:gan loss} is rewritten directly in terms of $p_G\left(x\right)$ and Eq. \ref{eq: true D*} rather than $\theta_G$ and $\theta_D^*\pp{\theta_G}$, then it is similarly a smooth function of $p_G\left(x\right)$. 
These smoothness guarantees are typically lost when $D\left(x; \theta_D \right)$ and $G\pp{ z; \theta_G}$ are drawn from parametric families.
They nonetheless suggest that the true generator objective in Eq. \ref{eq:gan loss} will often be well behaved, and is a desirable target for direct optimization.

Explicitly solving for the optimal discriminator parameters $\theta_D^*\pp{\theta_G}$ for every update step of the generator $G$ is computationally infeasible for discriminators based on neural networks. Therefore this minimax optimization problem is typically solved by alternating gradient descent on $\theta_G$ and ascent on $\theta_D$.

The optimal solution $\theta^* = \{\theta_G^*, \theta_D^* \}$ is a fixed point of these iterative learning dynamics. Additionally, if $f\pp{\theta_G, \theta_D}$ is convex in $\theta_G$ and concave in $\theta_D$, then alternating gradient descent (ascent) trust region updates are guaranteed to converge to the fixed point, under certain additional weak assumptions 
\citep{juditsky2011first}.
However in practice $f\pp{\theta_G, \theta_D}$ is typically very far from convex in $\theta_G$ and concave in $\theta_D$, and updates are not constrained in an appropriate way. As a result GAN training suffers from mode collapse, undamped oscillations, and other problems detailed in Section \ref{sec intro gan}. In order to address these difficulties, we will introduce a surrogate objective function $f_K\pp{\theta_G, \theta_D}$ for training the generator which more closely resembles the true generator objective $f\pp{\theta_G, \theta_D^*\pp{\theta_G}}$.

\subsection{Unrolling GANs}
\label{sec: unrolling}

A local optimum of the discriminator parameters $\theta_D^{*}$ can be expressed as the fixed point of an iterative optimization procedure,
\begin {align}
\theta_D^0 &= \theta_D 
\label{eq:unroll init}\\
\theta_D^{k+1} &= \theta_D^k + \eta^k \od{f\pp{\theta_G, \theta_D^k}}{\theta_D^k} \label{eq:sgd unroll}\\
\theta_D^*\left(\theta_G\right) &= \lim_{k \to \infty}\theta_D^k
,
\label{eq:unroll converge}
\end{align}
where $\eta^k$ is the learning rate schedule. For clarity, we have expressed Eq.~\ref{eq:sgd unroll} as a full batch steepest gradient ascent equation. More sophisticated optimizers can be similarly unrolled. In our experiments we unroll Adam \citep{Kingma14}.

By unrolling for $K$ steps, we create a surrogate objective for the update of the generator,
\begin {align}
\label{eq: surrogate loss}
f_K\pp{\theta_G, \theta_D} &= f\pp{\theta_G, \theta^K_D\pp{\theta_G, \theta_D}}
.
\end{align}
When $K=0$ this objective corresponds exactly to the standard GAN objective, while as $K\to\infty$ it corresponds to the true generator objective function $f\pp{\theta_G, \theta_D^*\pp{G}}$. By adjusting the number of unrolling steps $K$, we are thus able to interpolate between standard GAN training dynamics with their associated pathologies, and more costly gradient descent on the true generator loss.

\subsection{Parameter Updates}

\begin{figure}
  \includegraphics[width=\linewidth]{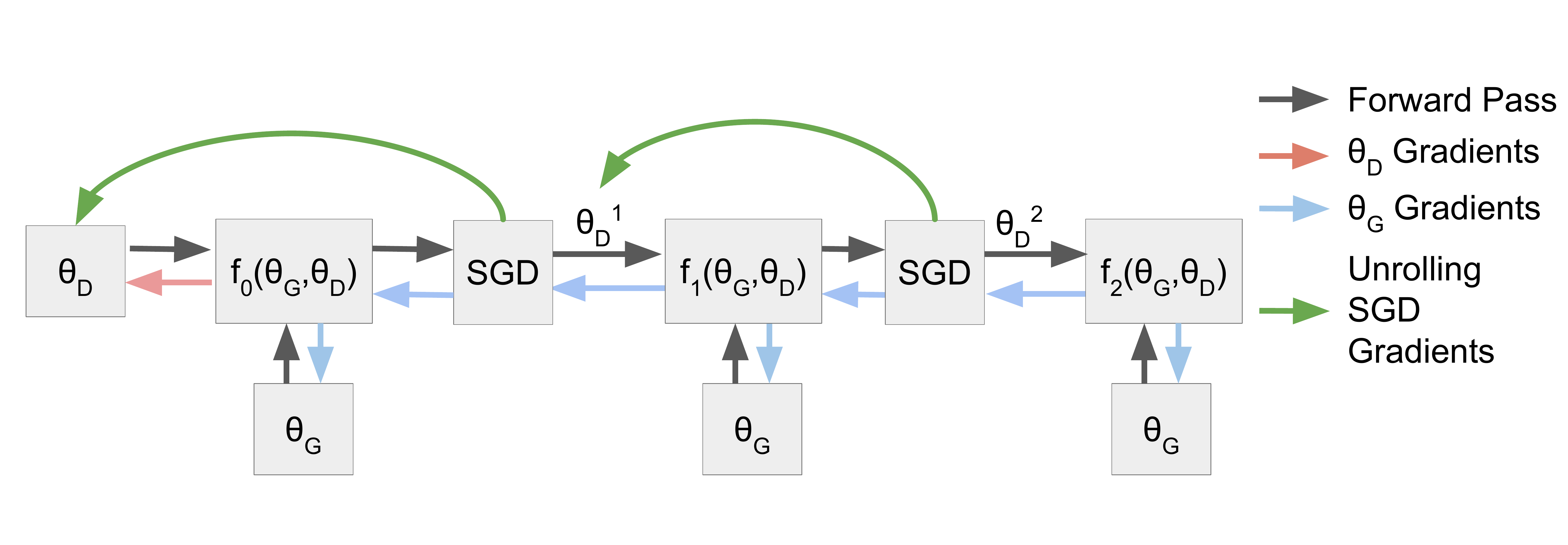}
  \caption{An illustration of the computation graph for an unrolled GAN with 3 unrolling steps.
  The generator update in Equation \ref{eq:G update} involves backpropagating the generator gradient (blue arrows) through the unrolled optimization. 
  Each step $k$ in the unrolled optimization uses the gradients of $f_k$ with respect to $\theta_D^k$, as described in Equation \ref{eq:sgd unroll} and indicated by the green arrows. 
  The discriminator update in Equation \ref{eq:D update} does not depend on the unrolled optimization (red arrow). 
  \label{fig:network}
 }
\end{figure}

The generator and discriminator parameter updates using this surrogate loss are
\begin {align}
\theta_G &\leftarrow \theta_G - \eta \od{f_K\pp{\theta_G, \theta_D}}{\theta_G} \label{eq:G update}\\
\theta_D &\leftarrow \theta_D + \eta \od{f\pp{\theta_G, \theta_D}}{\theta_D} \label{eq:D update}
.
\end{align}
For clarity we use full batch steepest gradient descent (ascent) with stepsize $\eta$ above, while in experiments we instead use minibatch Adam for both updates. The gradient in Eq.~\ref{eq:G update} requires backpropagating through the optimization process in Eq.~\ref{eq:sgd unroll}. A clear description of differentiation through gradient descent is given as Algorithm 2 in \citep{Maclaurin15}, though in practice the use of an automatic differentiation package means this step does not need to be programmed explicitly. 
A pictorial representation of these updates is provided in Figure \ref{fig:network}.

It is important to distinguish this from an approach suggested in \citep{Goodfellow14}, that several update steps of the discriminator parameters should be run before each single update step for the generator. In that approach, the update steps for both models are still gradient descent (ascent) with respect to {\em fixed} values of the other model parameters, rather than the surrogate loss we describe in Eq. \ref{eq: surrogate loss}.
Performing $K$ steps of discriminator update between each single step of generator update corresponds to updating the generator parameters $\theta_G$ using only the first term in Eq. \ref{eqn:optimal grad} below. 

\subsection{The Missing Gradient Term}

To better understand the behavior of the surrogate loss $f_K\pp{\theta_G, \theta_D}$, we examine its gradient with respect to the generator parameters $\theta_G$,
\begin{align} \label{eqn:optimal grad}
\od{f_K\pp{\theta_G, \theta_D}}{\theta_G} &=    
    \pd{f\pp{\theta_G, \theta_D^K\pp{\theta_G, \theta_D}}}{\theta_G}
    + 
    \pd{f\pp{\theta_G, \theta_D^K\pp{\theta_G, \theta_D}}}{\theta_D^K\pp{\theta_G, \theta_D}} \od{\theta_D^K\pp{\theta_G, \theta_D}}{\theta_G}
    . 
\end{align}

Standard GAN training corresponds exactly to updating the generator parameters using only the first term in this gradient, with $\theta_D^K\pp{\theta_G, \theta_D}$ being the parameters resulting from the discriminator update step.
An optimal generator for any fixed discriminator is a delta function at the $x$ to which the discriminator assigns highest data probability. Therefore, in standard GAN training, each generator update step is a partial collapse towards a delta function.

The second term captures how the discriminator would react to a change in the generator. It reduces the tendency of the generator to engage in mode collapse. For instance, the second term reflects that as the generator collapses towards a delta function, the discriminator reacts and assigns lower probability to that state, increasing the generator loss. It therefore discourages the generator from collapsing, and may improve stability.

As ${K\to\infty}$, $\theta^K_D$ goes to a local optimum of $f$, where $\pd{f}{\theta^K_D}=0$, and therefore the second term in Eq.~\ref{eqn:optimal grad} goes to 0 \citep{danskin2012theory}. 
The gradient of the unrolled surrogate loss $f_K\pp{\theta_G, \theta_D}$ with respect to $\theta_G$ is thus identical to the gradient of the standard GAN loss $f\pp{\theta_G, \theta_D}$ both when $K=0$ and when $K\to\infty$, where we take $K\to\infty$ to imply that in the standard GAN the discriminator is also fully optimized between each generator update. Between these two extremes, $f_K\pp{\theta_G, \theta_D}$ captures additional information about the response of the discriminator to changes in the generator.

\subsection{Consequences of the Surrogate Loss}
\label{gametheory}
GANs can be thought of as a game between the discriminator ($D$) and the generator ($G$). The agents take turns taking actions and updating their parameters until a Nash equilibrium is reached. The optimal action for $D$ is to evaluate the probability ratio $\frac{p_{data}(x)}{p_G(x) + p_{data}(x)}$ for the generator's move $x$ (Eq. \ref{eq: true D*}). The optimal generator action is to move its mass to maximize this ratio.

The initial move for $G$ will be to move as much mass as its parametric family and update step permits to the single point that maximizes the ratio of probability densities. The action $D$ will then take is quite simple. It will track that point, and to the extent allowed by its own parametric family and update step assign low data probability to it, and uniform probability everywhere else. This cycle of $G$ moving and $D$ following will repeat forever or converge depending on the rate of change of the two agents. This is similar to the situation in simple matrix games like rock-paper-scissors and matching pennies, where alternating gradient descent (ascent) with a fixed learning rate is known not to converge \citep{Singh2000, Bowling2002}.

In the unrolled case, however, this undesirable behavior no longer occurs. Now $G$'s actions take into account how $D$ will respond. In particular, $G$ will try to make steps that $D$ will have a hard time responding to. This extra information helps the generator spread its mass to make the next $D$ step less effective instead of collapsing to a point.

In principle, a surrogate loss function could be used for both $D$ and $G$. In the case of 1-step unrolled optimization this is known to lead to convergence for games in which gradient descent (ascent) fails \citep{Zhang10}. However, the motivation for using the surrogate generator loss in Section \ref{sec: unrolling}, of unrolling the {\em inner} of two nested $\min$ and $\max$ functions, does not apply to using a surrogate discriminator loss. Additionally, it is more common for the discriminator to overpower the generator than vice-versa when training a GAN. Giving more information to $G$ by allowing it to `see into the future' may thus help the two models be more balanced.

\section{Experiments}
\label{Sec:exp}
In this section we demonstrate improved mode coverage and stability by applying this technique to five datasets of increasing complexity. 
Evaluation of generative models is a notoriously hard problem \citep{Theis2016a}. As such the de facto standard in GAN literature has become sample quality as evaluated by a human and/or evaluated by a heuristic (Inception score for example, \citep{Salimans16}). 
While these evaluation metrics do a reasonable job capturing sample quality, they fail to capture sample diversity. In our first 2 experiments diversity is easily evaluated via visual inspection. In our later experiments this is not the case, and we will use a variety of methods to quantify coverage of samples. 
Our measures are individually strongly suggestive of unrolling reducing mode-collapse and improving stability, but none of them alone are conclusive. We believe that taken together however, they provide extremely compelling evidence for the advantages of unrolling. 

When doing stochastic optimization, we must choose which minibatches to use in the unrolling updates in Eq.~\ref{eq:sgd unroll}. 
We experimented with both a fixed minibatch and re-sampled minibatches for each unrolling step, and found it did not significantly impact the result. We use fixed minibatches for all experiments in this section.

We provide a reference implementation of this technique at \href{https://github.com/poolio/unrolled_gan}{github.com/poolio/unrolled\_gan}.

\subsection{Mixture of Gaussians Dataset}
\label{Sec:exp:toy}
\begin{figure}
  \includegraphics[width=\linewidth]{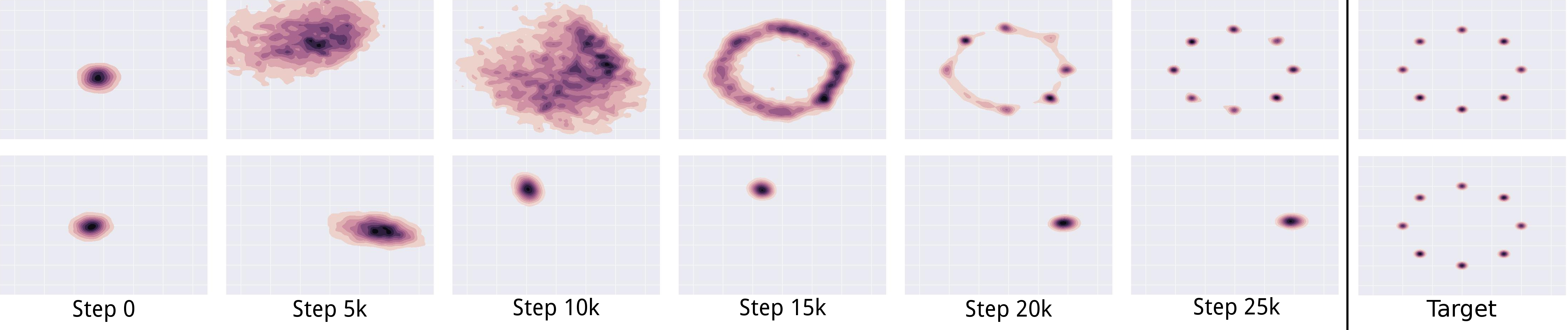}
  \caption{
  Unrolling the discriminator stabilizes GAN training on a toy 2D mixture of Gaussians dataset. Columns show a heatmap of the generator distribution after increasing numbers of training steps. The final column shows the data distribution.
  The top row shows training for a GAN with 10 unrolling steps.
  Its generator quickly spreads out and converges to the target distribution.
  The bottom row shows standard GAN training.
  The generator rotates through the modes of the data distribution. It never converges to a fixed distribution, and only ever assigns significant probability mass to a single data mode at once.
  \label{fig:toy}\label{exp:toy}}
\end{figure}

To illustrate the impact of discriminator unrolling, we train a simple GAN architecture on a 2D mixture of 8 Gaussians arranged in a circle. For a detailed list of architecture and hyperparameters see Appendix \ref{app gauss details}. Figure \ref{exp:toy} shows the dynamics of this model through time. Without unrolling the generator rotates around the valid modes of the data distribution but is never able to spread out mass. 
When adding in unrolling steps G quickly learns to spread probability mass and the system converges to the data distribution.

In Appendix \ref{app mix gauss} we perform further experiments on this toy dataset. We explore how unrolling compares to historical averaging, and compares to using the unrolled discriminator to update the generator, but without backpropagating through the generator. In both cases we find that the unrolled objective performs better.

\subsection{Pathological model with mismatched generator and discriminator}
\label{Sec:exp:path}
\begin{figure}
  \includegraphics[width=\linewidth]{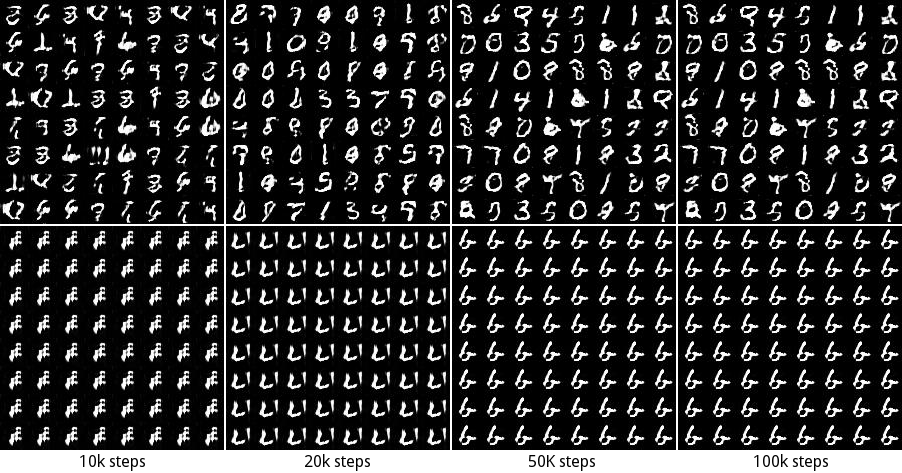}
  \caption{Unrolled GAN training increases stability for an RNN generator and convolutional discriminator trained on  MNIST. The top row was run with 20 unrolling steps. The bottom row is a standard GAN, with 0 unrolling steps. Images are samples from the generator after the indicated number of training steps.}
  \label{fig:rnn}
\end{figure}

To evaluate the ability of this approach to improve trainability, we look to a traditionally challenging family of models to train -- recurrent neural networks (RNNs). In this experiment we try to generate MNIST samples using an LSTM \citep{Hochreiter97}.
MNIST digits are 28x28 pixel images. At each timestep of the generator LSTM, it outputs one column of this image, so that after 28 timesteps it has output the entire sample. 
We use a convolutional neural network as the discriminator. See Appendix \ref{app:rnn_details} for the full model and training details. Unlike in all previously successful GAN models, there is no symmetry between the generator and the discriminator in this task, resulting in a more complex power balance. Results can be seen in Figure \ref{fig:rnn}. Once again, without unrolling the model quickly collapses, and rotates through a sequence of single modes. Instead of rotating spatially, it cycles through proto-digit like blobs. When running with unrolling steps the generator disperses and appears to cover the whole data distribution, as in the 2D example.

\subsection{Mode and manifold collapse using augmented MNIST}
GANs suffer from two different types of model collapse -- collapse to a subset of data modes, and collapse to a sub-manifold within the data distribution. 
In these experiments we isolate both effects using artificially constructed datasets, and demonstrate that unrolling can largely rescue both types of collapse.

\subsubsection{Discrete mode collapse}

To explore the degree to which GANs drop discrete modes in a dataset, we use a technique similar to one from \citep{Che16}.
We construct a dataset by stacking three randomly chosen MNIST digits, so as to construct an RGB image with a different MNIST digit in each color channel.
This new dataset has 1,000 distinct modes, corresponding to each combination of the ten MNIST classes in the three channels. 

We train a GAN on this dataset, and generate samples from the trained model (25,600 samples for all experiments). 
We then compute the predicted class label of each color channel using a pre-trained MNIST classifier. 
To evaluate performance, we use two metrics: the number of modes for which the generator produced at least one sample, and the KL divergence between the model and the expected data distribution. 
Within this discrete label space, a KL divergence can be estimated tractably between the generated samples and the data distribution over classes, where the data distribution is a uniform distribution over all 1,000 classes. 

As presented in Table \ref{exp:mnist3_table}, as the number of unrolling steps is increased, both mode coverage and reverse KL divergence improve.
Contrary to \citep{Che16}, we found that reasonably sized models (such as the one used in Section \ref{Sec:exp:img}) covered all 1,000 modes even without unrolling. 
As such we use smaller convolutional GAN models. 
Details on the models used are provided in Appendix \ref{app:1000 mnist}.

We observe an additional interesting effect in this experiment. The benefits of unrolling increase as the discriminator size is reduced. We believe unrolling effectively increases the capacity of the discriminator. The unrolled discriminator can better react to any specific way in which the generator is producing non-data-like samples. 
When the discriminator is weak, the positive impact of unrolling is thus larger.

\begin{table}
\centering
\makebox[\textwidth][c]{
\begin{tabular}{ |c |c | c | c| c | c |}
    \hline
    Discriminator Size & Unrolling steps & 0 & 1 & 5 & 10 \\
    \hline
    1/4 size of D compared to G & Modes generated & 30.6 $\pm$ 20.73 & 65.4 $\pm$ 34.75 & 236.4 $\pm$ 63.30 & \textbf{327.2 $\pm$ 74.67} \\
    \hline
    & KL(model \textbar \textbar data) & 5.99 $\pm$ 0.42 & 5.911 $\pm$ 0.14 & 4.67 $\pm$ 0.43 & \textbf{4.66 $\pm$ 0.46} \\
    \hline
    1/2 size of D compared to G & Modes generated & 628.0 $\pm$ 140.9 & 523.6 $\pm$ 55.768 & 732.0 $\pm$ 44.98 & \textbf{817.4 $\pm$ 37.91} \\
    \hline
    & KL(model \textbar \textbar data) & 2.58 $\pm$0.751 &2.44 $\pm$0.26 & 1.66 $\pm$ 0.090 & \textbf{1.43 $\pm$ 0.12} \\
    \hline
\end{tabular}
}
\caption{Unrolled GANs cover more discrete modes when modeling a dataset with 1,000 data modes, corresponding to all combinations of three MNIST digits ($10^3$ digit combinations). The number of modes covered is given for different numbers of unrolling steps, and for two different architectures. The reverse KL divergence between model and data is also given. Standard error is provided for both measures.}
\label{exp:mnist3_table}
\end{table}

\subsubsection{Manifold collapse}

In addition to discrete modes, we examine the effect of unrolling when modeling continuous manifolds. 
To get at this quantity, we constructed a dataset consisting of colored MNIST digits. 
Unlike in the previous experiment, a single MNIST digit was chosen, and then assigned a single monochromatic color. 
With a perfect generator, one should be able to recover the distribution of colors used to generate the digits. 
We use colored MNIST digits so that the generator also has to model the digits, which makes the task sufficiently  complex that the generator is unable to perfectly solve it. 
The color of each digit is sampled from a 3D normal distribution. Details of this dataset are provided in Appendix \ref{app:color mnist}. 
We will examine the distribution of colors in the samples generated by the trained GAN. 
As will also be true in the CIFAR10 example in Section \ref{Sec:exp:img}, the lack of diversity in generated colors is almost invisible using only visual inspection of the samples. Samples can be found in Appendix \ref{app:color mnist}.

In order to recover the color the GAN assigned to the digit, we used k-means with 2 clusters, to pick out the foreground color from the background.
We then performed this transformation for both the training data and the generated images.
Next we fit a Gaussian kernel density estimator to both distributions over digit colors. Finally, we computed the JS divergence between the model and data distributions over colors.
Results can be found in Table \ref{exp:mnist_color_table} for several model sizes. Details of the models are provided in Appendix \ref{app:color mnist}.

In general, the best performing models are unrolled for 5-10 steps, and larger models perform better than smaller models. Counter-intuitively, taking 1 unrolling step seems to hurt this measure of diversity. We suspect that this is due to it introducing oscillatory dynamics into training. Taking more unrolling steps however leads to improved performance with unrolling.

\begin{table}
\centering
\makebox[\textwidth][c]{
\begin{tabular}{ |c | c | c| c | c |}
    \hline
    Unrolling steps & 0 & 1 & 5 & 10 \\
    \hline
    JS divergence with 1/4 layer size & 0.073 $\pm$ 0.0058 & 0.142 $\pm$ 0.028 & \textbf{0.049 $\pm$ 0.0021} & 0.075 $\pm$ 0.012 \\
    \hline
    JS divergence with 1/2 layer size & 0.095 $\pm$ 0.011 & 0.119 $\pm$ 0.010 & \textbf{0.055 $\pm$ 0.0049} & 0.074$\pm$ 0.016 \\
    \hline
    JS divergence with 1/1 layer size & 0.034 $\pm$ 0.0034 & 0.050 $\pm$ 0.0026 &0.027 $\pm$ 0.0028 & \textbf{0.025 $\pm$ 0.00076} \\
    \hline

\end{tabular}
}
\caption{Unrolled GANs better model a continuous distribution. GANs are trained to model randomly colored MNIST digits, where the color is drawn from a Gaussian distribution. The JS divergence between the data and model distributions over digit colors is then reported, along with standard error in the JS divergence. More unrolling steps, and larger models, lead to better JS divergence.
}
\label{exp:mnist_color_table}
\end{table}

\subsection{Image modeling of CIFAR10}
\label{Sec:exp:img}
\begin{figure}
\centering
\begin{tabular}{cc}
\hspace*{-.8cm} 
    \includegraphics[width=\linewidth]{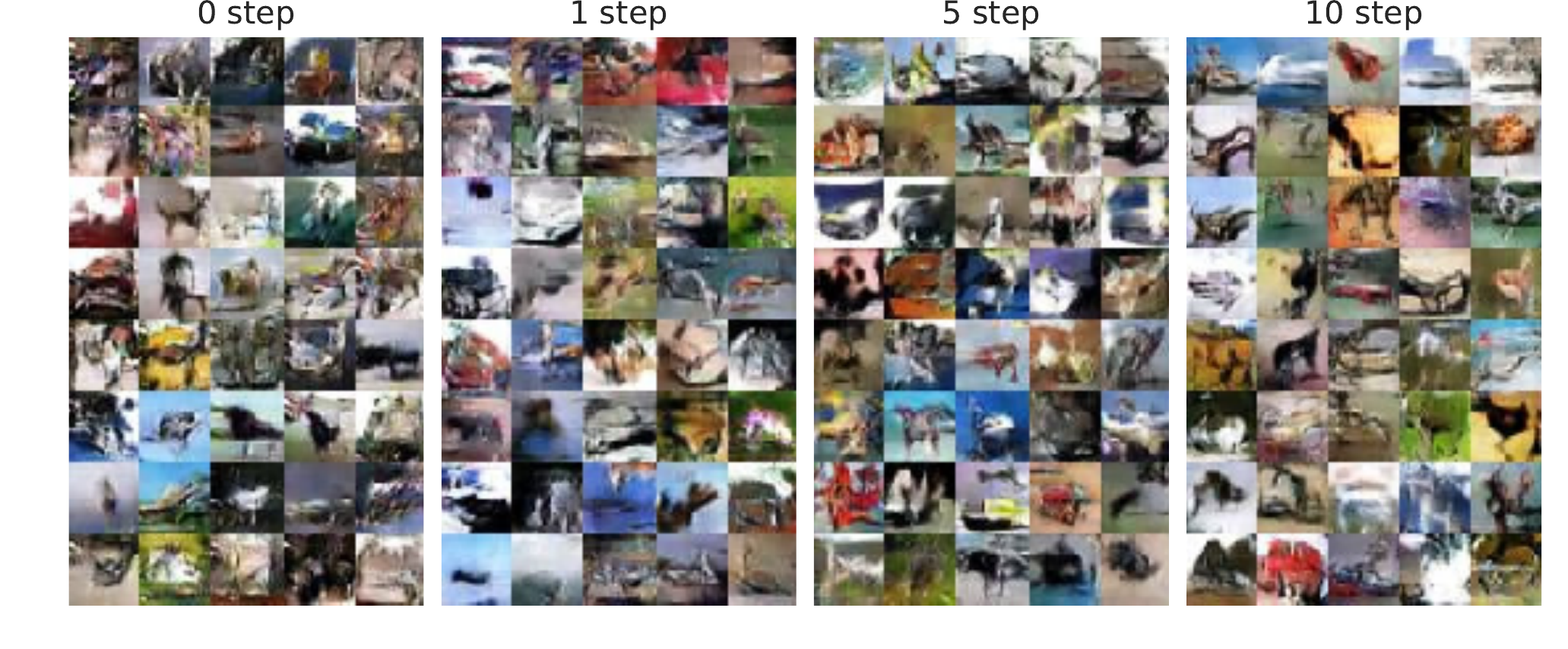}
\end{tabular}
\caption{Visual perception of sample quality and diversity is very similar for models trained with different numbers of unrolling steps. Actual sample diversity is higher with more unrolling steps. Each pane shows samples generated after training a model on CIFAR10 with 0, 1, 5, and 10 steps of unrolling.
}
\label{exp:image_cifar}
\end{figure}

Here we test our technique on a more traditional convolutional GAN architecture and task, similar to those used in \citep{Radford15, Salimans16}. 
In the previous experiments we tested models where the standard GAN training algorithm would not converge. 
In this section we improve a standard model by reducing its tendency to engage in mode collapse.
We ran 4 configurations of this model, varying the number of unrolling steps to be 0, 1, 5, or 10. Each configuration was run 5 times with different random seeds. For full training details see Appendix \ref{app:conv}. Samples from each of the 4 configurations can be found in Figure \ref{exp:image_cifar}. There is no obvious difference in visual quality across these model configurations. Visual inspection however provides only a poor measure of sample diversity.

By training with an unrolled discriminator, we expect to generate more diverse samples which more closely resemble the underlying data distribution.
We introduce two techniques to examine sample diversity: inference via optimization, and pairwise distance distributions.

\subsubsection{Inference via Optimization}
Since likelihood cannot be tractably computed, over-fitting of GANs is typically tested by taking samples and computing the nearest-neighbor images in pixel space from the training data \citep{Goodfellow14}. 
We will do the reverse, and measure the ability of the generative model to generate images that look like specific samples from the training data.
If we did this by generating random samples from the model, we would need an exponentially large number of samples.
We instead treat finding the nearest neighbor $x_\text{nearest}$ to a target image $x_\text{target}$ as an optimization task,
\begin{align}
z_\text{nearest} &= \argmin_{z}\norm{G\pp{z; \theta_G} - x_\text{target}}_2^2 \\
x_\text{nearest} &= G\pp{z_\text{nearest}; \theta_G}.
\end{align}

This concept of backpropagating to generate images has been widely used in visualizing features from discriminative networks \citep{Simonyan2013, Yosinski2015, Nguyen2016} and has been applied to explore the visual manifold of GANs in \citep{zhu2016generative}.

We apply this technique to each of the models trained. We optimize with 3 random starts using LBFGS, which is the optimizer typically used in similar settings such as style transfer \citep{Johnson16,Champandard16}. 
Results comparing average mean squared errors between $x_\text{nearest}$ and $x_\text{target}$ in pixel space can be found in Table \ref{exp:table_cifar}. 
In addition we compute the percent of images for which a certain configuration achieves the lowest loss when compared to the other configurations. 

In the zero step case, there is poor reconstruction and less than 1\% of the time does it obtain the lowest error of the 4 configurations. 
Taking 1 unrolling step results in a significant improvement in MSE. Taking 10 unrolling steps results in more modest improvement, but continues to reduce the reconstruction MSE.

\begin{figure}
\centering
\makebox[\textwidth][c]{
\includegraphics[width=0.52\linewidth]{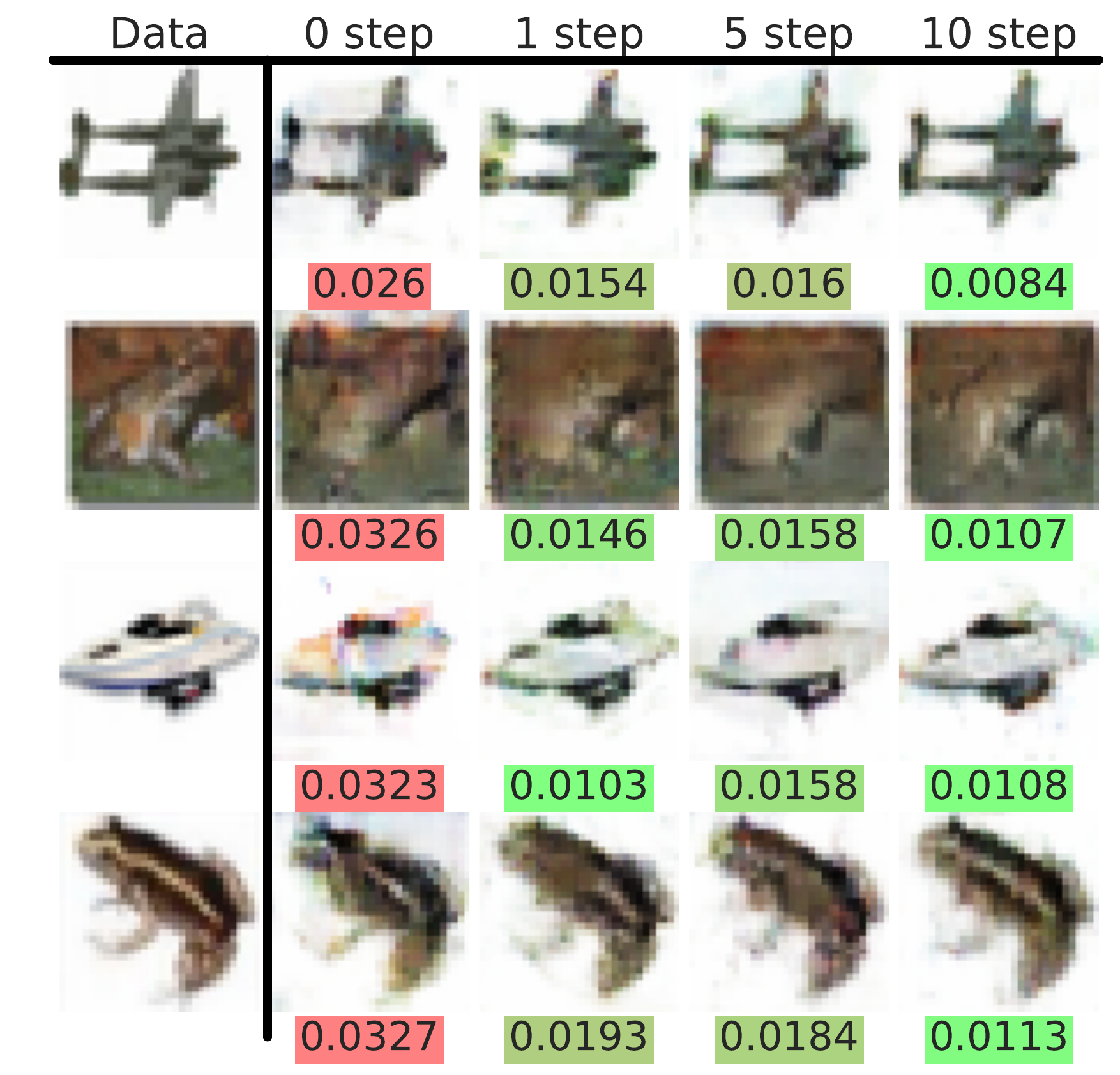}
\includegraphics[width=0.52\linewidth]{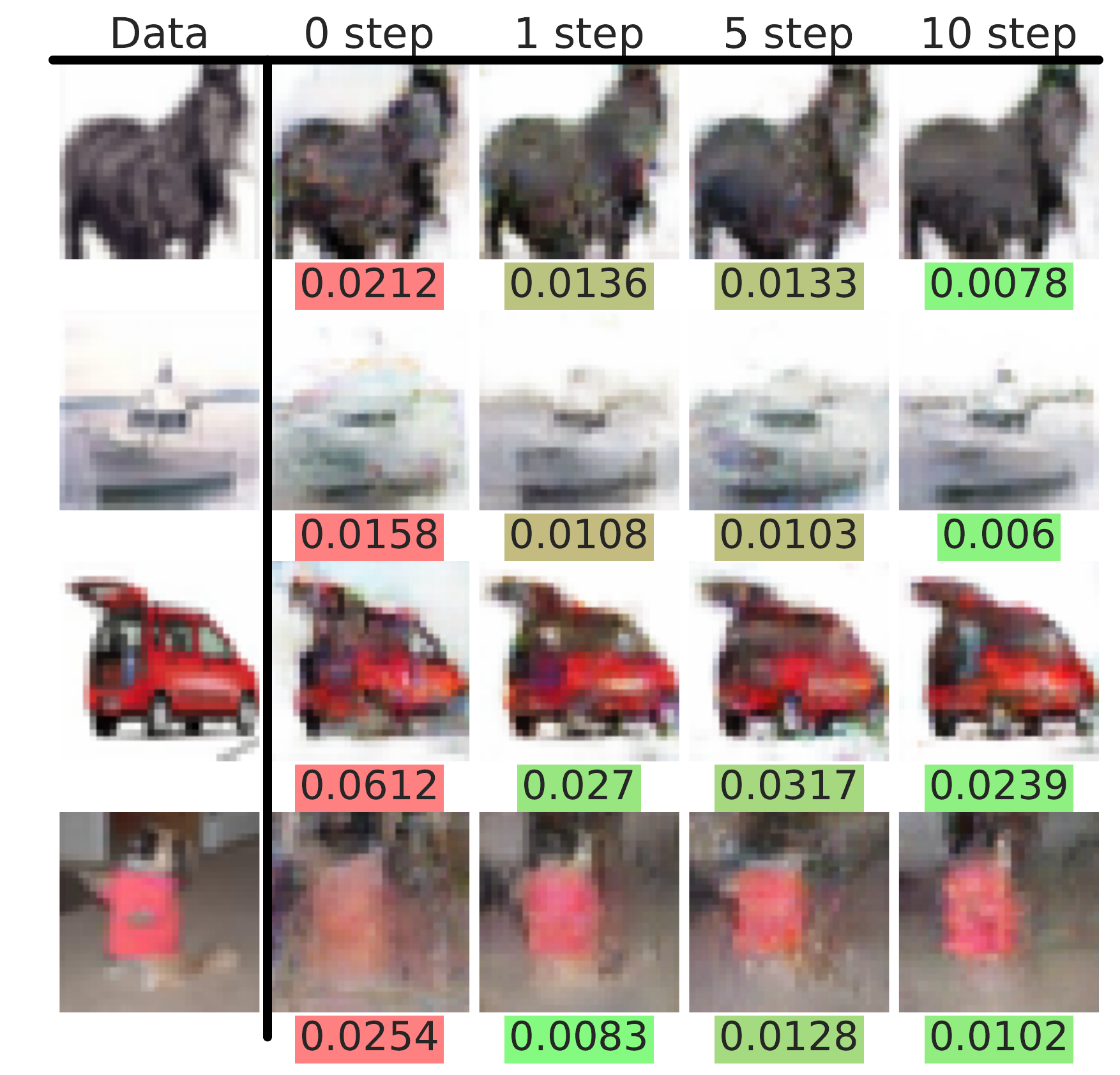}
}
\caption{
Training set images are more accurately reconstructed using GANs trained with unrolling than by a standard (0 step) GAN, likely due to mode dropping by the standard GAN.
Raw data is on the left, and the optimized images to reach this target follow for 0, 1, 5, and 10 unrolling steps. 
The reconstruction MSE is listed below each sample. 
A random 1280 images where selected from the training set, and corresponding best reconstructions for each model were found via optimization. 
Shown here are the eight images with the largest absolute fractional difference between GANs trained with 0 and 10 unrolling steps.
}
\label{exp:optim}
\end{figure}

To visually see this, we compare the result of the optimization process for 0, 1, 5, and 10 step configurations in Figure \ref{exp:optim}. 
To select for images where differences in behavior is most apparent, we sort the data by the absolute value of a fractional difference in MSE between the 0 and 10 step models, 
$\left|\frac{l_{0step}-l_{10step}}{\frac{1}{2}\left(l_{0step}+l_{10step}\right)}\right|$. 
This highlights examples where either the 0 or 10 step model cannot accurately fit the data example but the other can. 
In Appendix \ref{app:more_images} we show the same comparison for models initialized using different random seeds. 
Many of the zero step images are fuzzy and ill-defined suggesting that these images cannot be generated by the standard GAN generative model, and come from a dropped mode. 
As more unrolling steps are added, the outlines become more clear and well defined -- the model covers more of the distribution and thus can recreate these samples.

\begin{table}
\centering
\centering
\makebox[\textwidth][c]{
\begin{tabular}{ |c | c | c| c | c |}
    \hline
  Unrolling Steps & 0 steps & 1 step & 5 steps & 10 steps \\
  \hline			
  Average MSE & 0.0231 $\pm$ 0.0024 & 0.0195 $\pm$  0.0021  & 0.0200 $\pm$ 0.0023 & \textbf{0.0181 $\pm$  0.0018} \\
  \hline
  Percent Best Rank & 0.63\% & 22.97\% & 15.31\% & \textbf{61.09\%} \\
  \hline
\end{tabular}
}

\caption{
GANs trained with unrolling are better able to match images in the training set than standard GANs, likely due to mode dropping by the standard GAN. 
Results show the MSE between training images and the best reconstruction for a model with the given number of unrolling steps. 
The fraction of training images best reconstructed by a given model is given in the final column. 
The best reconstructions is found by optimizing the latent representation $z$ to produce the closest matching pixel output $G\pp{z; \theta_G}$. 
Results are averaged over all 5 runs of each model with different random seeds.
\label{exp:table_cifar}
}
\end{table}

\subsubsection{Pairwise Distances}
\begin{figure}
\centering
\includegraphics[width=.8\linewidth]{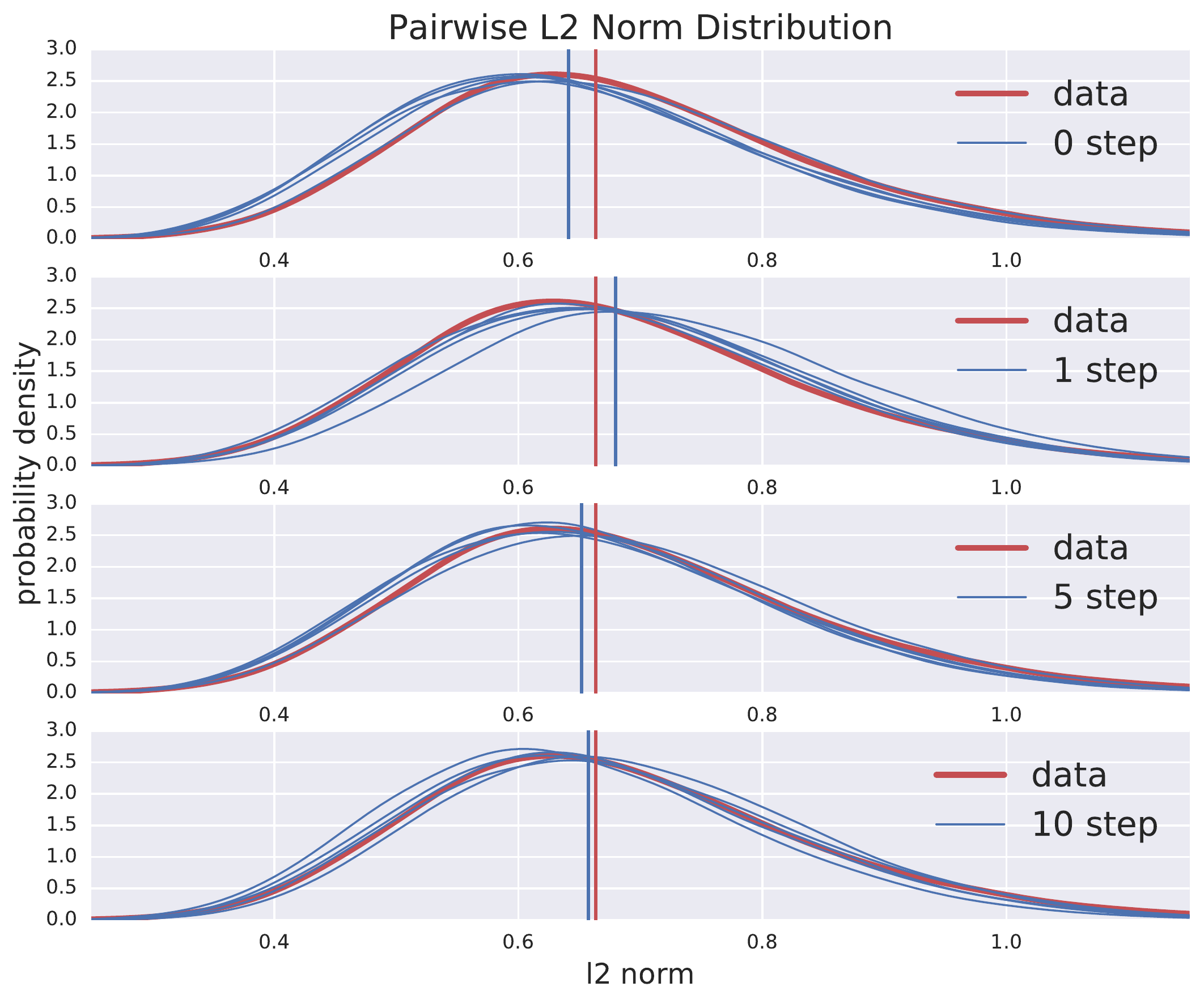}
\caption{As the number of unrolling steps in GAN training is increased, the distribution of pairwise distances between model samples more closely resembles the same distribution for the data.
Here we plot histograms of pairwise distances between randomly selected samples. The red line gives pairwise distances in the data,
while each of the five blue lines in each plot represents a model trained with a different random seed.
The vertical lines are the medians of each distribution.}\label{exp:image_prob}

\end{figure}

A second complementary approach is to compare statistics of data samples to the corresponding statistics for samples generated by the various models. 
One particularly simple and relevant statistic is the distribution over pairwise distances between random pairs of samples. 
In the case of mode collapse, greater probability mass will be concentrated in smaller volumes, and the distribution over inter-sample distances should be skewed towards smaller distances.
We sample random pairs of images from each model, as well as from the training data, and compute histograms of the $\ell_2$ distances between those sample pairs.
As illustrated in Figure \ref{exp:image_prob}, the standard GAN, with zero unrolling steps, has its probability mass skewed towards smaller $\ell_2$ intersample distances, compared to real data. 
As the number of unrolling steps is increased, the histograms over intersample distances increasingly come to resemble that for the data distribution. 
This is further evidence in support of unrolling decreasing the mode collapse behavior of GANs.

\section{Discussion}

In this work we developed a method to stabilize GAN training and reduce mode collapse by defining the generator objective with respect to unrolled optimization of the  discriminator. 
We then demonstrated the application of this method to several tasks, where it either rescued unstable training, or reduced the tendency of the model to drop regions of the data distribution.

The main drawback to this method is computational cost of each training step, which increases linearly with the number of unrolling steps. There is a tradeoff between better approximating the true generator loss and the computation required to make this estimate. Depending on the architecture, one unrolling step can be enough. In other more unstable models, such as the RNN case, more are needed to stabilize training. We have some initial positive results suggesting it may be sufficient to further perturb the training gradient in the same direction that a single unrolling step perturbs it. While this is more computationally efficient, further investigation is required.

The method presented here bridges some of the gap between theoretical and practical results for training of GANs. We believe developing better update rules for the generator and discriminator is an important line of work for GAN training. In this work we have only considered a small fraction of the design space. For instance, the approach could be extended to unroll $G$ when updating $D$ as well -- letting the discriminator react to how the generator would move. It is also possible to unroll sequences of $G$ and $D$ updates. This would make updates that are recursive: $G$ could react to maximize performance as if $G$ and $D$ had already updated.

\subsubsection*{Acknowledgments}
We would like to thank Laurent Dinh, David Dohan, Vincent Dumoulin, Liam Fedus, Ishaan Gulrajani, Julian Ibarz, Eric Jang, Matthew Johnson, Marc Lanctot, Augustus Odena, Gabriel Pereyra, Colin Raffel, Sam Schoenholz, Ayush Sekhari, Jon Shlens, and Dale Schuurmans for insightful conversation, as well as the rest of the Google Brain Team.

\newpage
\bibliography{iclr2017_conference}
\bibliographystyle{iclr2017_conference}

\newpage
\appendix
\part*{Appendix}

\setcounter{figure}{0} \renewcommand{\thefigure}{App.\arabic{figure}}
\setcounter{table}{0} \renewcommand{\thetable}{App.\arabic{table}}

\section{2D Gaussian Training Details}\label{app gauss details}
Network architecture and experimental details for the experiment in Section~\ref{Sec:exp:toy} are as follows:

The dataset is sampled from a mixture of 8 Gaussians of standard deviation 0.02. The means are equally spaced around a circle of radius 2.

The generator network consists of a fully connected network with 2 hidden layers of size 128 with relu activations followed by a linear projection to 2 dimensions. All weights are initialized to be orthogonal with scaling of 0.8.

The discriminator network first scales its input down by a factor of 4 (to roughly scale to (-1,1)), followed by 1 layer fully connected network with relu activations to a linear layer to of size 1 to act as the logit.

The generator minimizes $\mathcal{L}_G = \mathrm{log}(D(x)) + \mathrm{log}(1-D(G(z)))$ and the discriminator minimizes $\mathcal{L}_D = - \mathrm{log}(D(x)) - \mathrm{log}(1-D(G(z)))$  where x is sampled from the data distribution and $z \sim \mathcal{N}(0, I_{256})$. Both networks are optimized using Adam \citep{Kingma14} with a learning rate of 1e-4 and $\beta_1$=0.5.

The network is trained by alternating updates of the generator and the discriminator. One step consists of either G or D updating.
 \label{app:toy_details}
 
\section{More Mixture of Gaussian Experiments}\label{app mix gauss}
\subsection{Effects of time delay / historical averaging}
Another comparison we looked at was with regard to historical averaging based approaches. Recently similarly inspired approaches have been used in \citep{Salimans16} to stabilize training. For our study, we looked at taking an ensemble of discriminators over time.

\begin{figure}[h]
\centering
\includegraphics[width=1.0\linewidth]{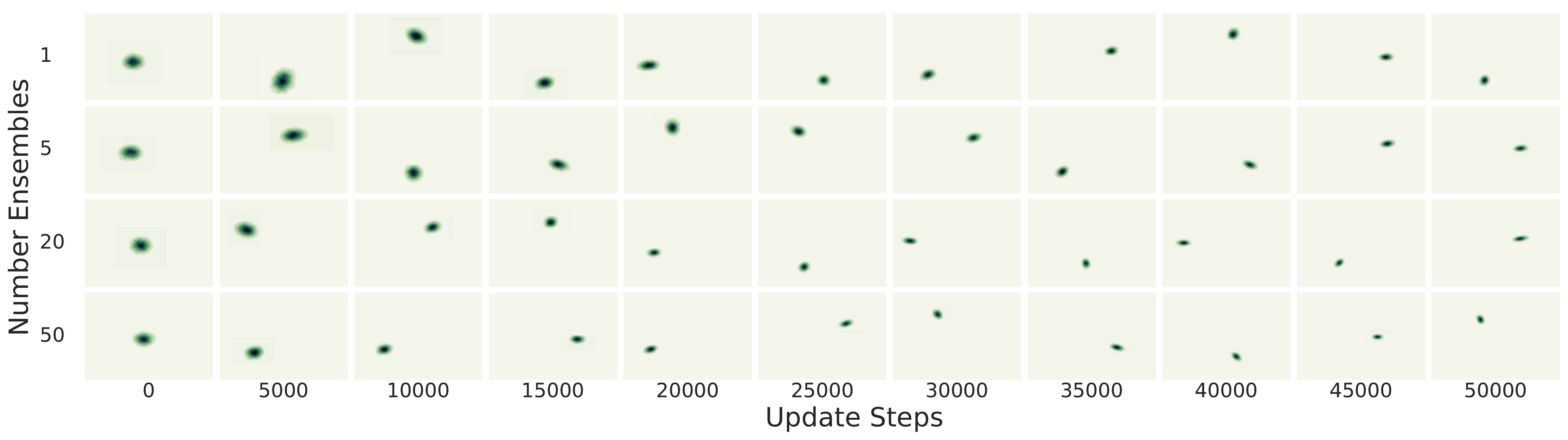}
\caption{Historical averaging does not visibly increase stability on the mixture of Gaussians task.
Each row corresponds to an ensemble of discriminators which consists of the indicated number of immediately preceding discriminators. 
The columns correspond to different numbers of training steps.}
 \label{exp:history_every}
\end{figure}

First, we looked at taking an ensemble of the last N steps, as shown in Figure \ref{exp:history_every}.

\begin{figure}[h]
\centering
\includegraphics[width=1.0\linewidth]{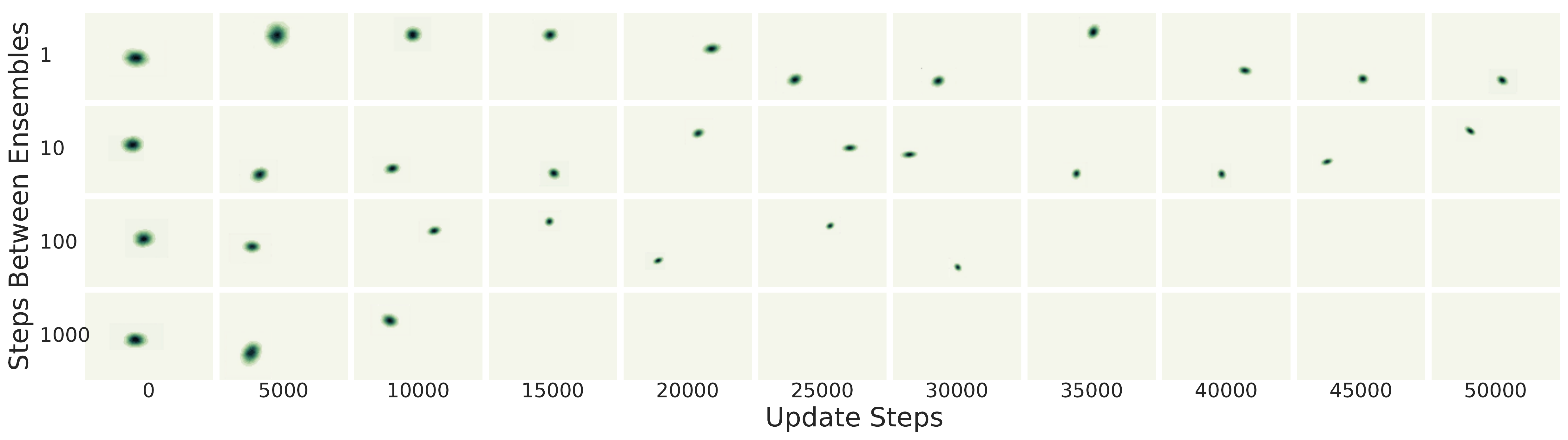}
\caption{Introducing longer time delays between the discriminator ensemble results in instability and probability distributions that are not in the window being visualized. The x axis is the number of weight updates and the y axis is how many steps to skip between discriminator updates when selecting the ensemble of 5 discriminators. }
 \label{exp:history_skip}
\end{figure}

To further explore this idea, we ran experiments with an ensemble of 5 discriminators, but with different periods between replacing discriminators in the ensemble. For example, if I sample at a rate of 100, it would take 500 steps to replace all 5 discriminators. Results can be seen in Figure \ref{exp:history_skip}.

We observe that given longer and longer time delays, the model becomes less and less stable. We hypothesize that this is due to the initial shape of the discriminator loss surface. When training, the discriminator's estimates of probability densities are only accurate on regions where it was trained. When fixing this discriminator, we are removing the feedback between the generator exploitation and the discriminators ability to move. As a result, the generator is able to exploit these fixed areas of poor performance for older discriminators in the ensemble. New discriminators (over)compensate for this, leading the system to diverge.

\subsection{Effects of the second gradient}
\begin{figure}[h]
\centering
Unrolled GAN
\includegraphics[width=1.0\linewidth]{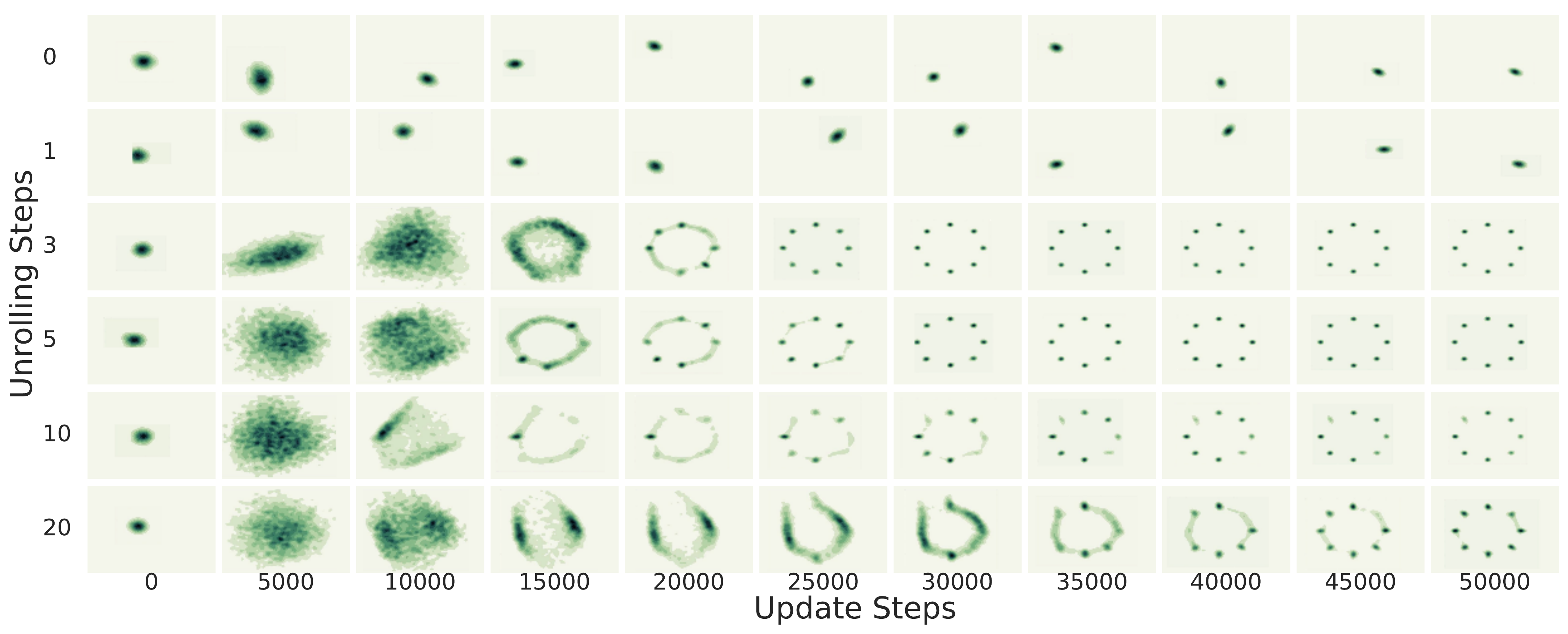}
Unrolled GAN without second gradient
\includegraphics[width=1.0\linewidth]{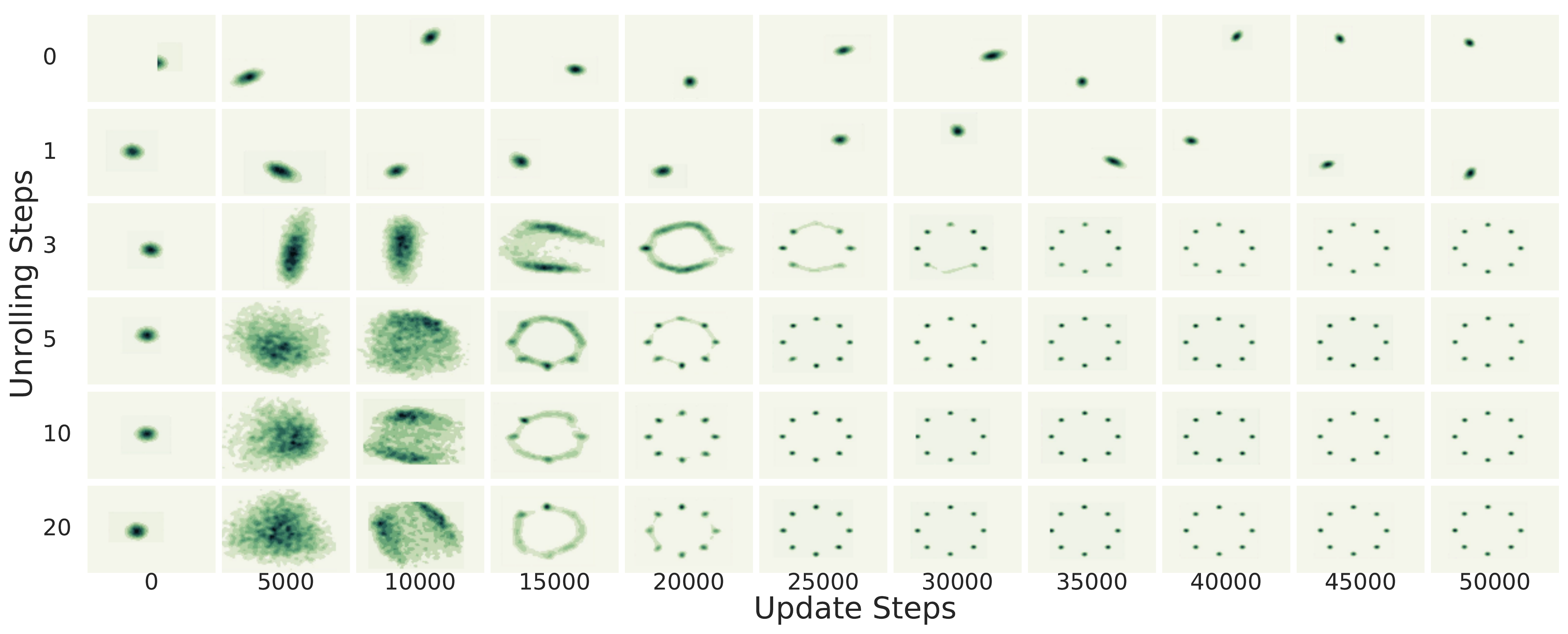}

\caption{If the discriminator remains nearly at its optimum during learning, then performance is nearly identical with and without the second gradient term in Equation \ref{eqn:optimal grad}. 
As shown in Figure \ref{exp:unroll_5g}, when the discriminator lags behind the generator, backpropagating through unrolling aids convergence.
\label{exp:unroll_1g}
}
\end{figure}

A second factor we analyzed is the effect of backpropagating the learning signal through the unrolling in Equation \ref{eqn:optimal grad}. 
We can turn on or off this backpropagation through the unrolling by introducing stop\_gradient calls into our computation graph between each unrolling step. 
With the stop\_gradient in place, the update signal corresponds only to the first term in Equation \ref{eqn:optimal grad}.
We looked at 3 configurations: without stop\_gradients; vanilla unrolled GAN, with stop gradients; and with stop gradients but taking the average over the $k$ unrolling steps instead of taking the final value. Results can be see in Figure \ref{exp:unroll_1g}.

We initially observed no difference between unrolling with and without the second gradient, as both required 3 unrolling steps to become stable. When the discriminator is unrolled to convergence, the second gradient term becomes zero. Due to the simplicity of the problem, we suspect that the discriminator nearly converged for every generator step, and the second gradient term was thus irrelevant.

\begin{figure}[h]
\centering
Unrolled GAN with 5 G Steps per D
\includegraphics[width=1.0\linewidth]{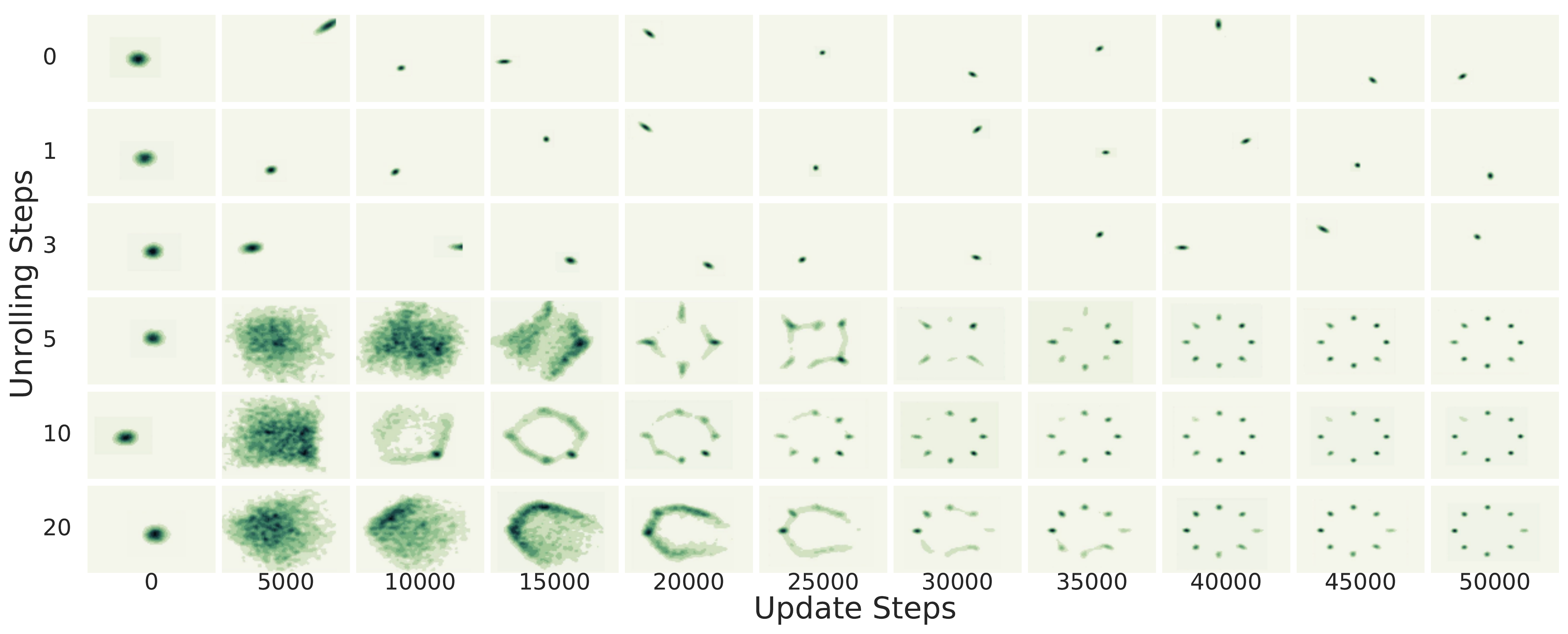}
Unrolled GAN with 5 G Steps per D without second gradient
\includegraphics[width=1.0\linewidth]{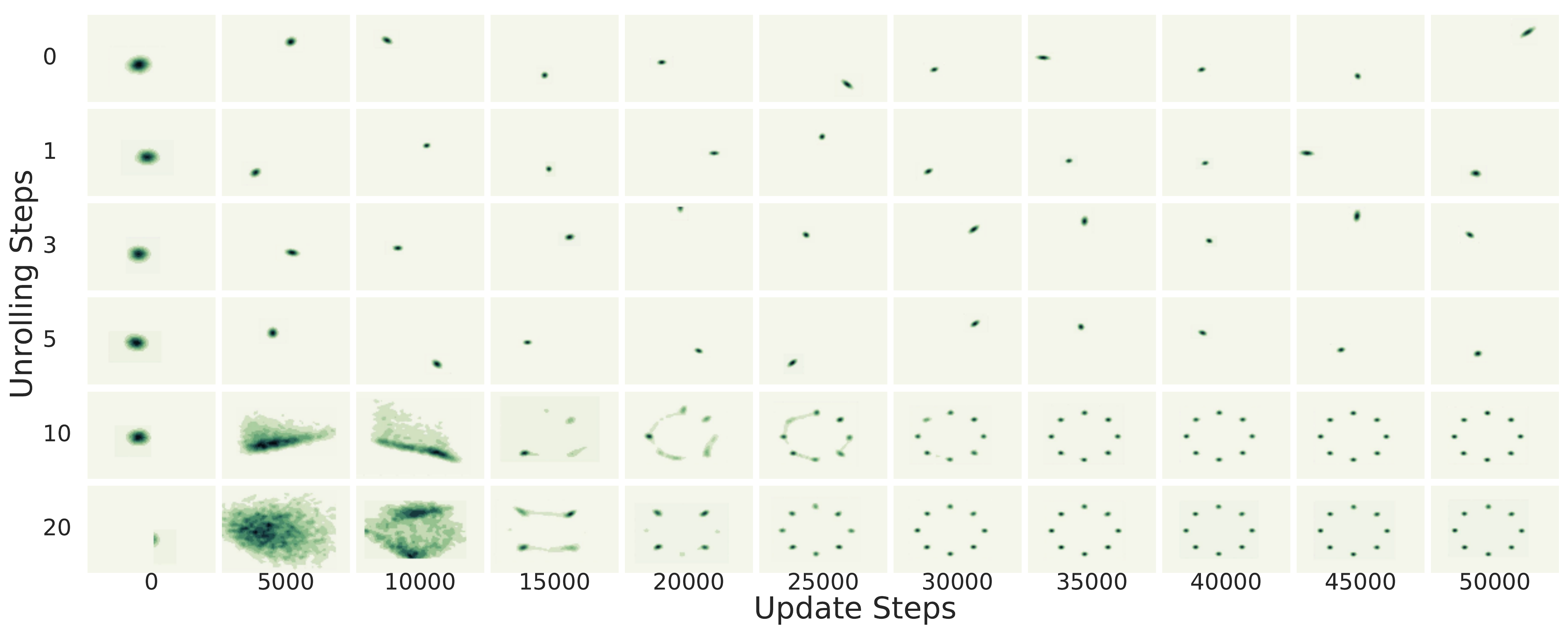}
\caption{
Backpropagating through the unrolling process aids convergence when the discriminator does not fully converge between generator updates. 
When taking 5 generator steps per discriminator step unrolling greatly increases stability, requiring only 5 unrolling steps to converge. Without the second gradient it requires 10 unrolling steps. 
Also see Figure \ref{exp:unroll_1g}.
\label{exp:unroll_5g}
}
\end{figure}

To test this, we modified the dynamics to perform five generator steps for each discriminator update. Results are shown in Figure \ref{exp:unroll_5g}. With the discriminator now kept out of equilibrium, successful training can be achieved with half as many unrolling steps when using both terms in the gradient than when only including the first term.

\section{RNN MNIST Training Details} \label{app:rnn_details}
The network architecture for the experiment in Section~\ref{Sec:exp:path} is as follows:

The MNIST dataset is scaled to [-1, 1).

The generator first scales the 256D noise vector through a 256 unit fully connected layer with relu activation. This is then fed into the initial state of a 256D LSTM\citep{Hochreiter97} that runs 28 steps corresponding to the number of columns in MNIST. The resulting sequence of activations is projected through a fully connected layer with 28 outputs with a tanh activation function. All weights are initialized via the "Xavier" initialization \citep{Glorot10}. The forget bias on the LSTM is initialized to 1.

The discriminator network feeds the input into a Convolution(16, stride=2) followed by a Convolution(32, stride=2) followed by Convolution(32, stride=2). All convolutions have stride 2. As in \citep{Radford15} leaky rectifiers are used with a 0.3 leak. Batch normalization is applied after each layer \citep{Ioffe15}. The resulting 4D tensor is then flattened and a linear projection is performed to a single scalar.

The generator network minimises $\mathcal{L}_G = \mathrm{log}(D(G(z)))$ and the discriminator minimizes $\mathcal{L}_D = \mathrm{log}(D(x)) + \mathrm{log}(1-D(G(z)))$. Both networks are trained with Adam\citep{Kingma14} with learning rates of 1e-4 and $\beta_1$=0.5. The network is trained alternating updating the generator and the discriminator for 150k steps. One step consists of just 1 network update.

\section{CIFAR10/MNIST Training Details} \label{app:conv}
The network architectures for the discriminator, generator, and encoder as as follows. All convolutions have a kernel size of 3x3 with batch normalization. The discriminator uses leaky ReLU's with a 0.3 leak and the generator uses standard ReLU.

The generator network is defined as:

\begin{tabular}{ l | l | l }
    \hline
       & number outputs & stride \\
    \hline
  Input: $z \sim \mathcal{N}(0, I_{256})$ & &\\
  Fully connected & 4 * 4 * 512 \\
  Reshape to image 4,4,512 & & \\
  Transposed Convolution & 256 & 2 \\
  Transposed Convolution & 128 & 2 \\
  Transposed Convolution & 64 & 2 \\
  Convolution & 1 or 3 & 1 \\
\end{tabular}

The discriminator network is defined as:

\begin{tabular}{ l | l | l }
    \hline
   & number outputs & stride \\
       \hline
  Input: $x \sim p_{data}$ or $G$ & &\\
  Convolution & 64 & 2 \\
  Convolution & 128 & 2 \\
  Convolution & 256 & 2 \\
  Flatten & & \\
  Fully Connected & 1 &\\
\end{tabular}

The generator network minimises $\mathcal{L}_G = \mathrm{log}(D(G(z)))$ and the discriminator minimizes $\mathcal{L}_D = \mathrm{log}(D(x)) + \mathrm{log}(1-D(G(z)))$. The networks are trained with Adam with a generator learning rate of 1e-4, and a discriminator learning rate of 2e-4. The network is trained alternating updating the generator and the discriminator for 100k steps. One step consists of just 1 network update.

\section{1000 class MNIST}\label{app:1000 mnist}

\begin{tabular}{ l | l | l }
    \hline
       & number outputs & stride \\
    \hline
  Input: $z \sim \mathcal{N}(0, I_{256})$ & &\\
  Fully connected & 4 * 4 * 64 \\
  Reshape to image 4,4,64 & & \\
  Transposed Convolution & 32 & 2 \\
  Transposed Convolution & 16 & 2 \\
  Transposed Convolution & 8 & 2 \\
  Convolution & 3 & 1 \\
\end{tabular}

The discriminator network is parametrized by a size X and is defined as follows. In our tests, we used X of 1/4 and 1/2.

\begin{tabular}{ l | l | l }
    \hline
   & number outputs & stride \\
       \hline
  Input: $x \sim p_{data}$ or $G$ & &\\
  Convolution & 8*X & 2 \\
  Convolution & 16*X & 2 \\
  Convolution & 32*X & 2 \\
  Flatten & & \\
  Fully Connected & 1 &\\
\end{tabular}

\section{Colored MNIST dataset}\label{app:color mnist}
\subsection{Dataset}
\begin{figure}[h]
\centering
\makebox[\textwidth][c]{
  \includegraphics[width=.30\linewidth]{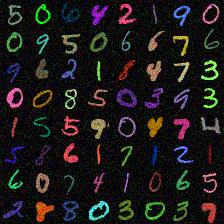}
  \includegraphics[width=.30\linewidth]{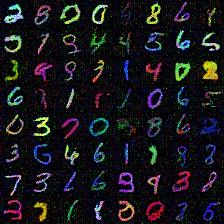}
  \includegraphics[width=.30\linewidth]{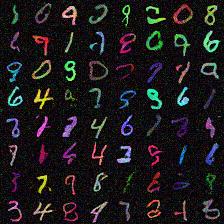}
  }

  \caption{Right: samples from the data distribution. Middle: Samples from 1/4 size model with 0 look ahead steps (worst diversity). Left: Samples from 1/1 size model with 10 look ahead steps (most diversity).\label{exp:mnist_color_samples}}  
\end{figure}

To generate this dataset we first took the mnist digit, $I$, scaled between 0 and 1. For each image we sample a color, $C$, normally distributed with mean=0 and std=0.5. To generate a colored digit between (-1, 1) we do $I*C + (I - 1)$. Finally, we add a small amount of pixel independent noise sampled from a normal distribution with std=0.2, and the resulting values are cliped between (-1, 1).  When visualized, this generates images and samples that can be seen in figure \ref{exp:mnist_color_samples}. Once again it is very hard to visually see differences in sample diversity when comparing the 128 and the 512 sized models.

\subsection{Models}
The models used in this section are parametrized by a variable X to control capacity. A value of X=1 is same architecture used in the cifar10 experiments. We used 1/4, 1/2 and 1 as these values.

The generator network is defined as:

\begin{tabular}{ l | l | l }
    \hline
       & number outputs & stride \\
    \hline
  Input: $z \sim \mathcal{N}(0, I_{256})$ & &\\
  Fully connected & 4 * 4 * 512*X \\
  Reshape to image 4,4,512*X & & \\
  Transposed Convolution & 256*X & 2 \\
  Transposed Convolution & 128*X & 2 \\
  Transposed Convolution & 64*X & 2 \\
  Convolution & 3 & 1 \\
\end{tabular}

The discriminator network is defined as:

\begin{tabular}{ l | l | l }
    \hline
   & number outputs & stride \\
       \hline
  Input: $x \sim p_{data}$ or $G$ & &\\
  Convolution & 64*X & 2 \\
  Convolution & 128*X & 2 \\
  Convolution & 256*X & 2 \\
  Flatten & & \\
  Fully Connected & 1 &\\
\end{tabular}

\section{Optimization Based Visualizations}
\label{app:more_images}
More examples of model based optimization. We performed 5 runs with different seeds of each of of the unrolling steps configuration. Bellow are comparisons for each run index. Ideally this would be a many to many comparison, but for space efficiency we grouped the runs by the index in which they were run. 

\begin{figure}[h]
\centering
\makebox[\textwidth][c]{
  \includegraphics[width=.50\linewidth]{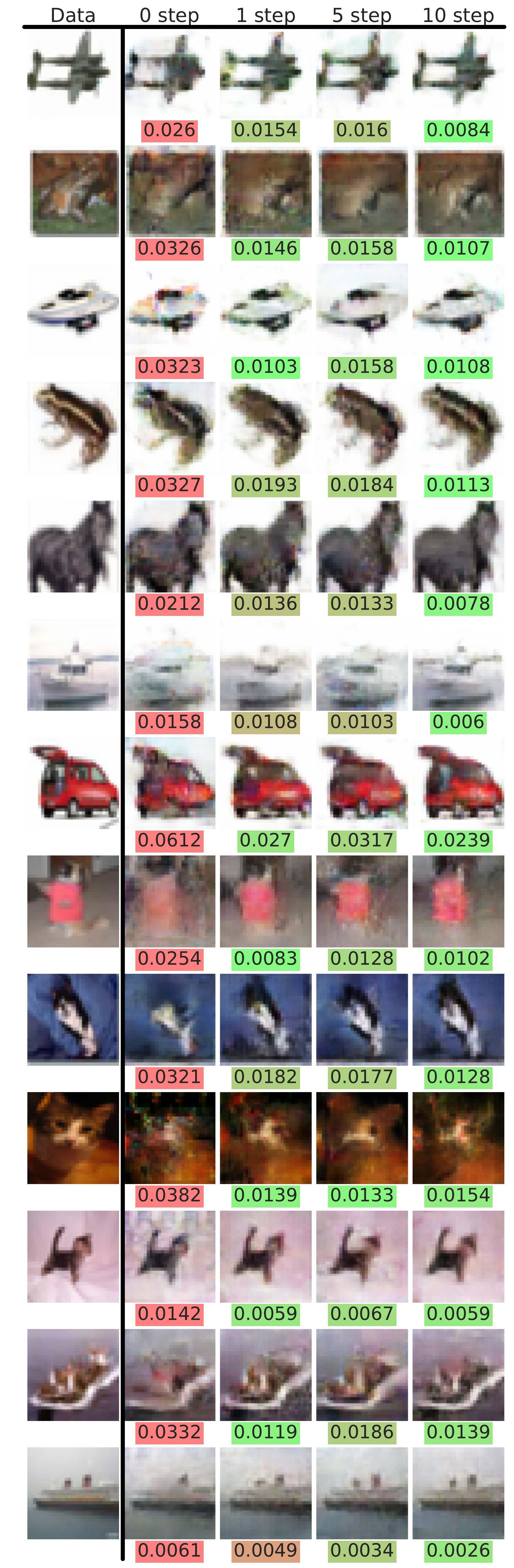}
  \includegraphics[width=.50\linewidth]{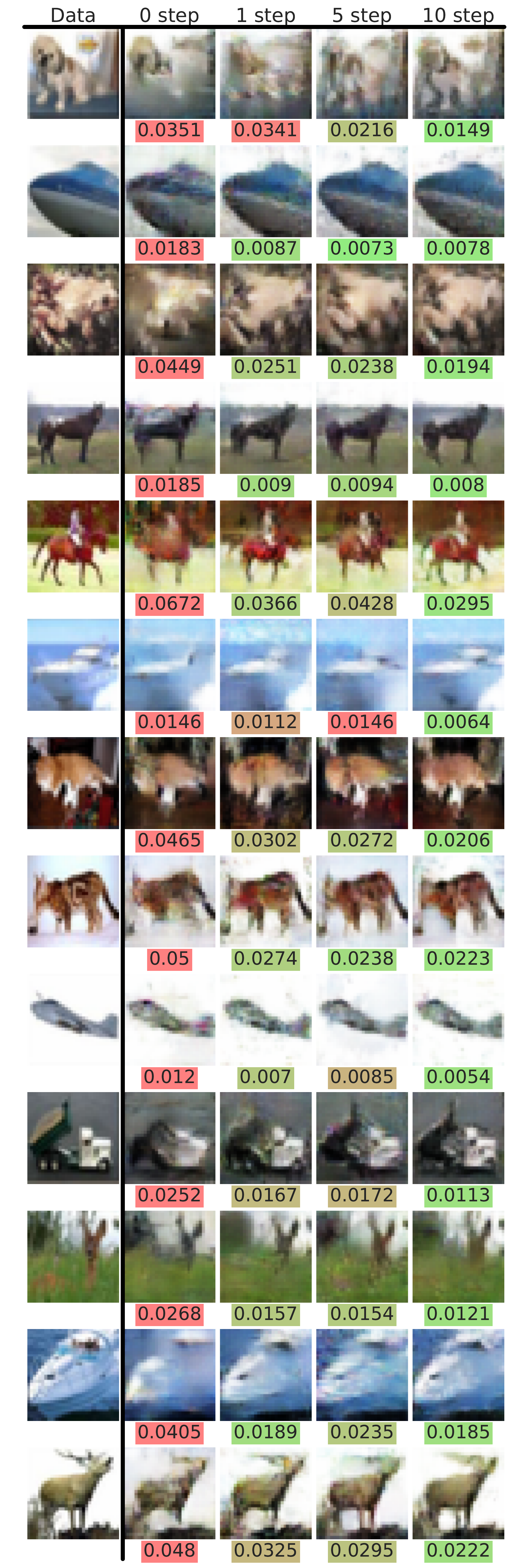}
  }
  \caption{Samples from 1/5 with different random seeds.}
\end{figure}

\begin{figure}[h]
\centering
\makebox[\textwidth][c]{
  \includegraphics[width=.50\linewidth]{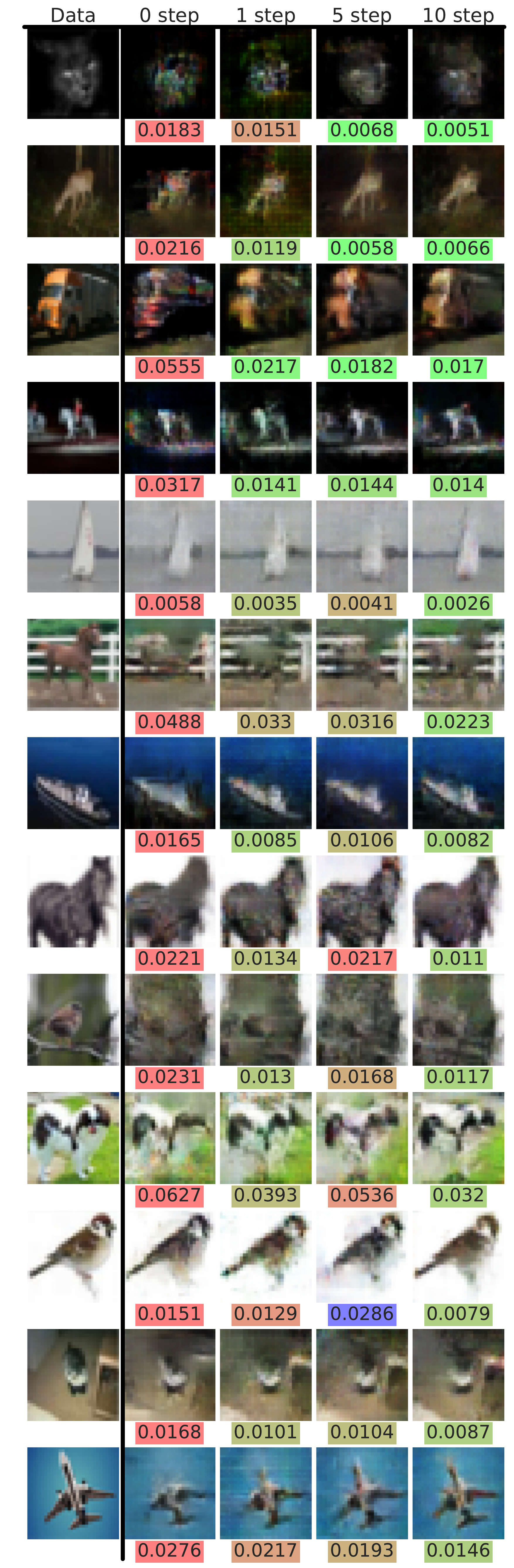}
  \includegraphics[width=.50\linewidth]{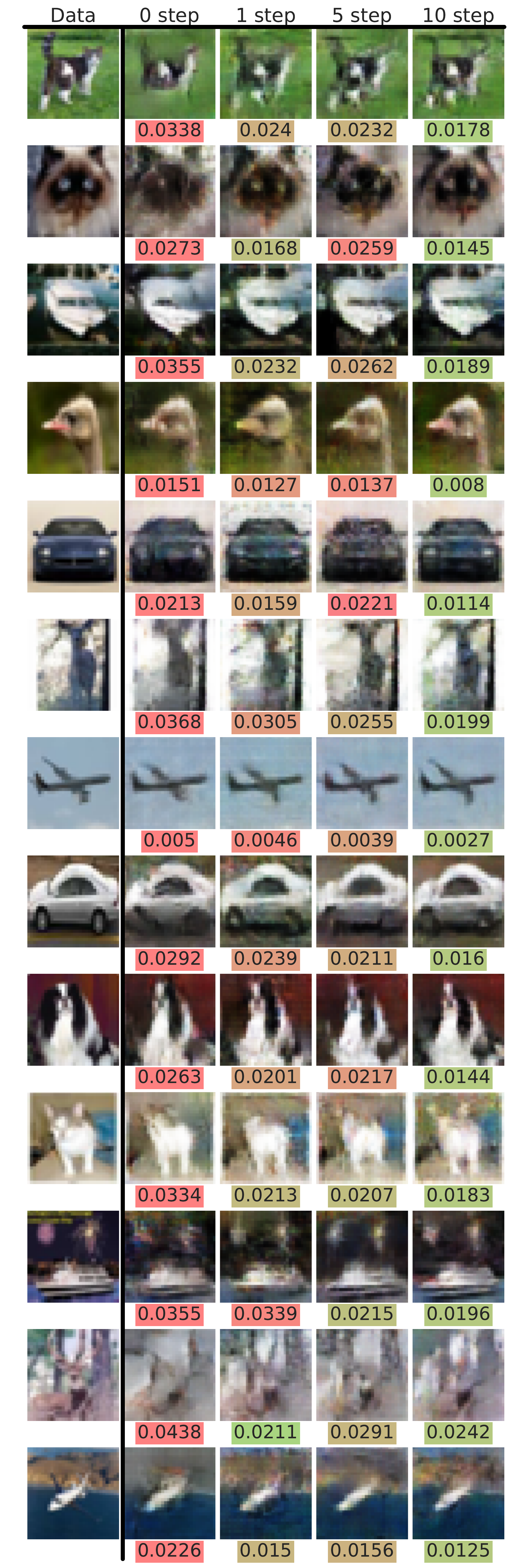}
  }
  \caption{Samples from 2/5 with different random seeds.}

\end{figure}

\begin{figure}[h]
\centering
\makebox[\textwidth][c]{
  \includegraphics[width=.50\linewidth]{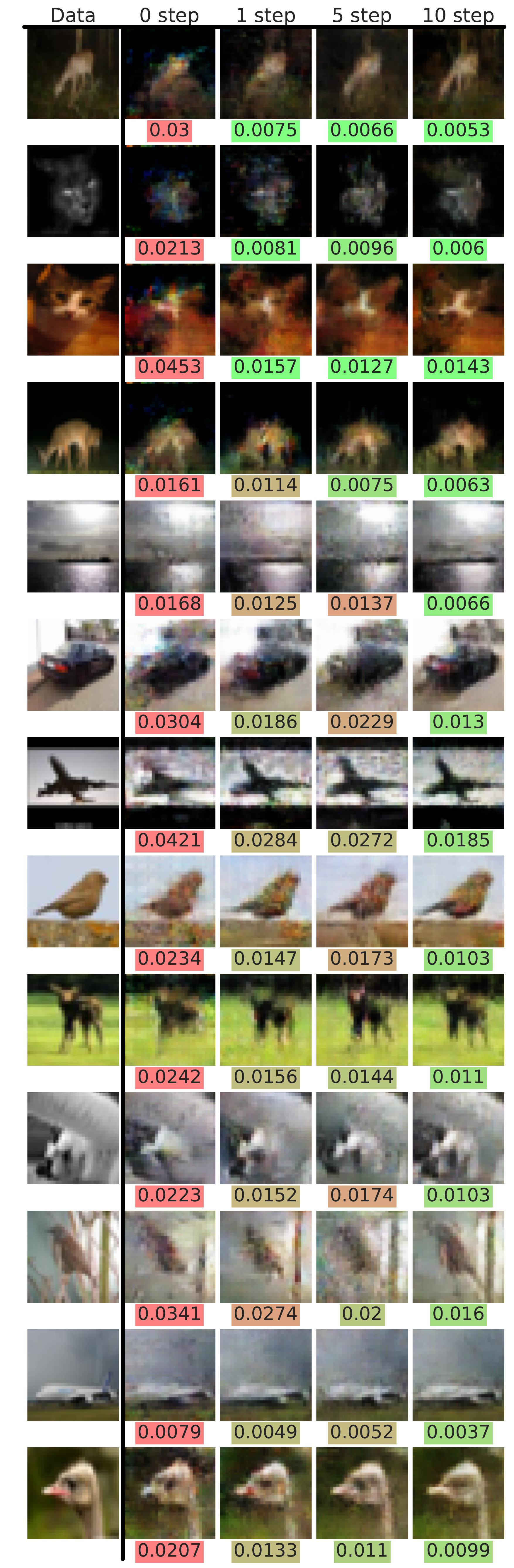}
  \includegraphics[width=.50\linewidth]{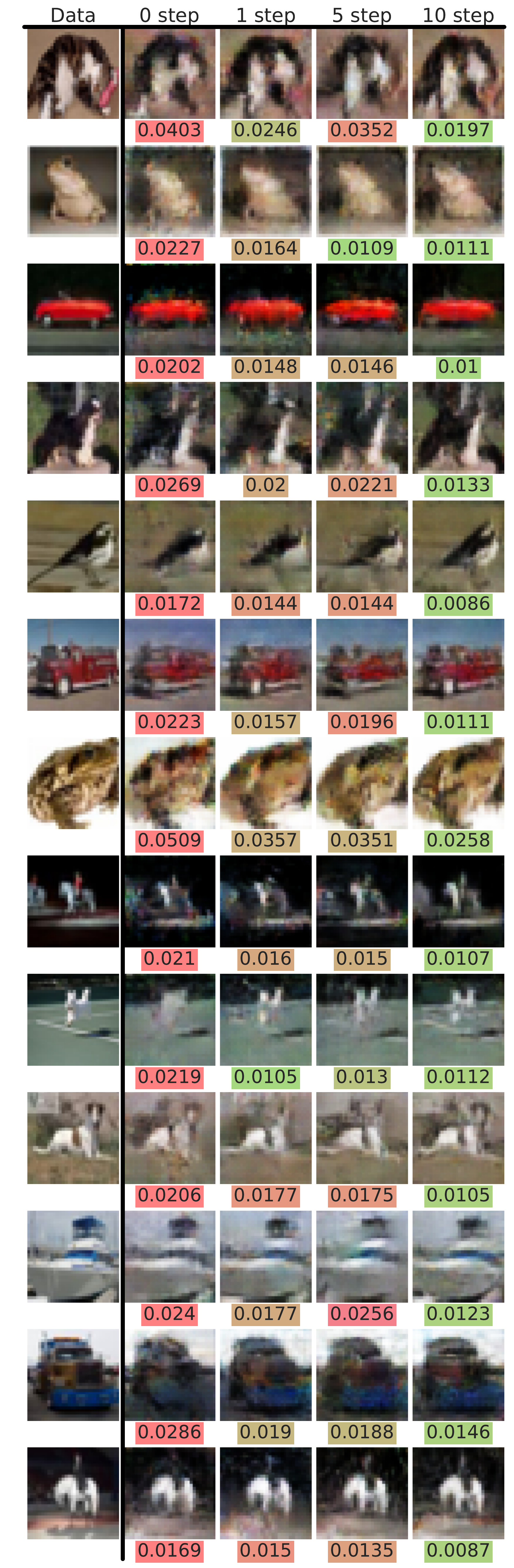}
  }
  \caption{Samples from 3/5 with different random seeds.}
\end{figure}

\begin{figure}[h]
\centering
\makebox[\textwidth][c]{
  \includegraphics[width=.50\linewidth]{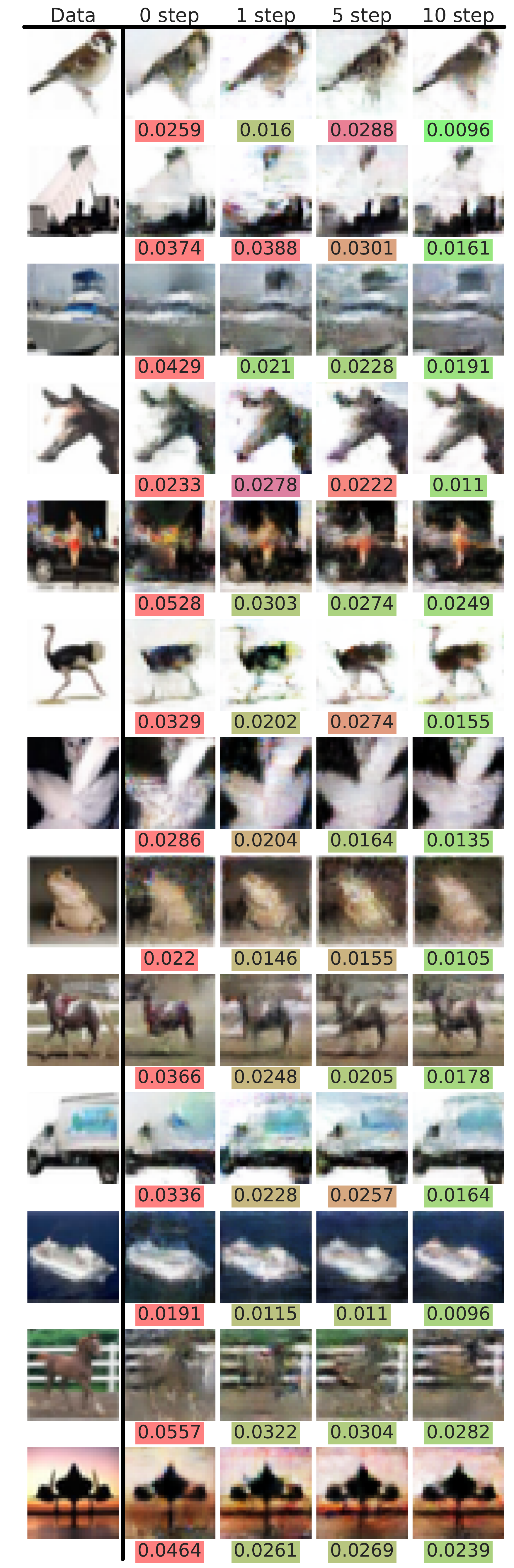}
  \includegraphics[width=.50\linewidth]{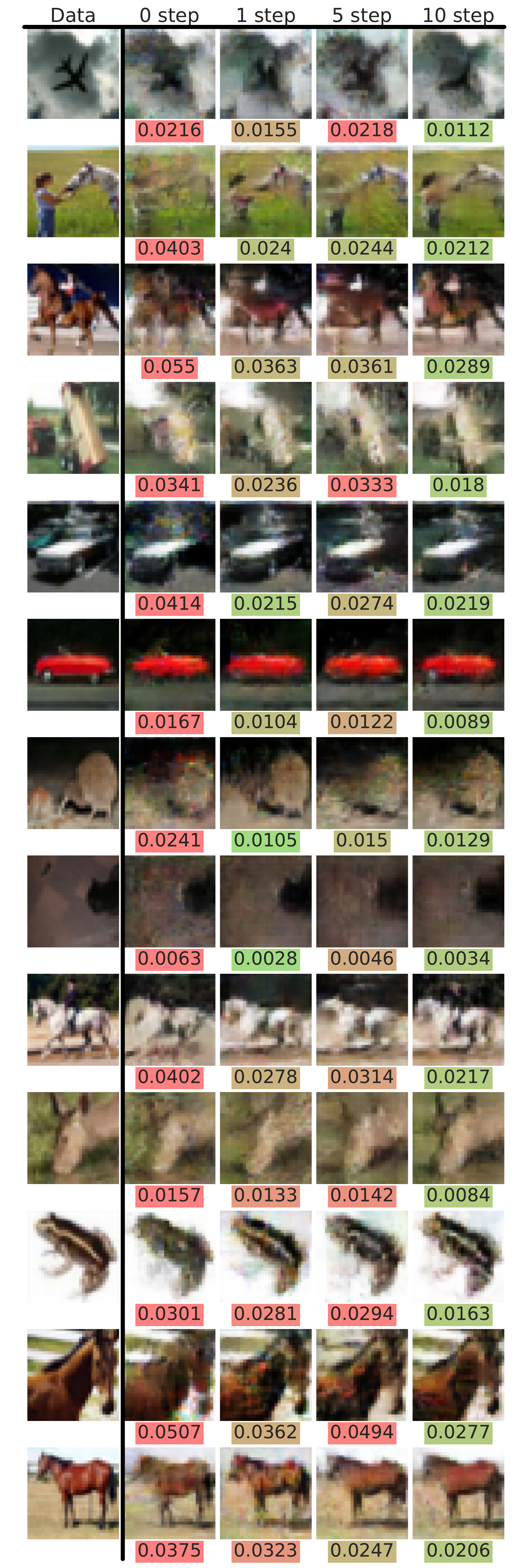}
  }
    \caption{Samples from 4/5 with different random seeds.}
\end{figure}

\begin{figure}[h]
\centering
\makebox[\textwidth][c]{
  \includegraphics[width=.50\linewidth]{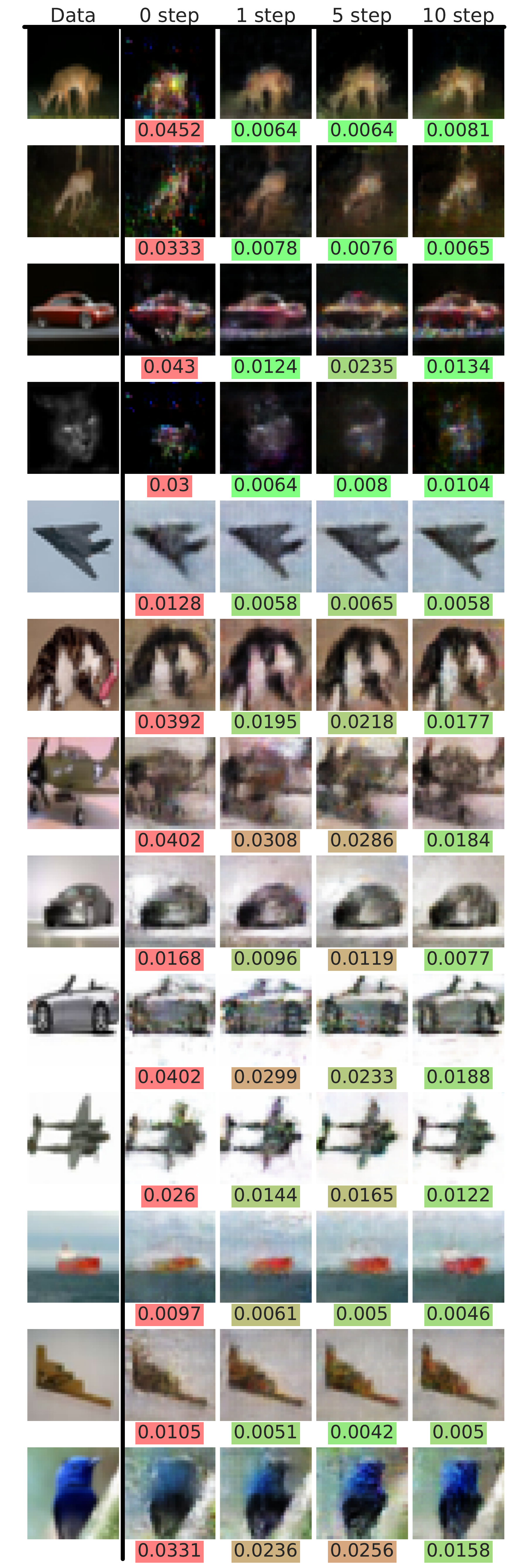}
  \includegraphics[width=.50\linewidth]{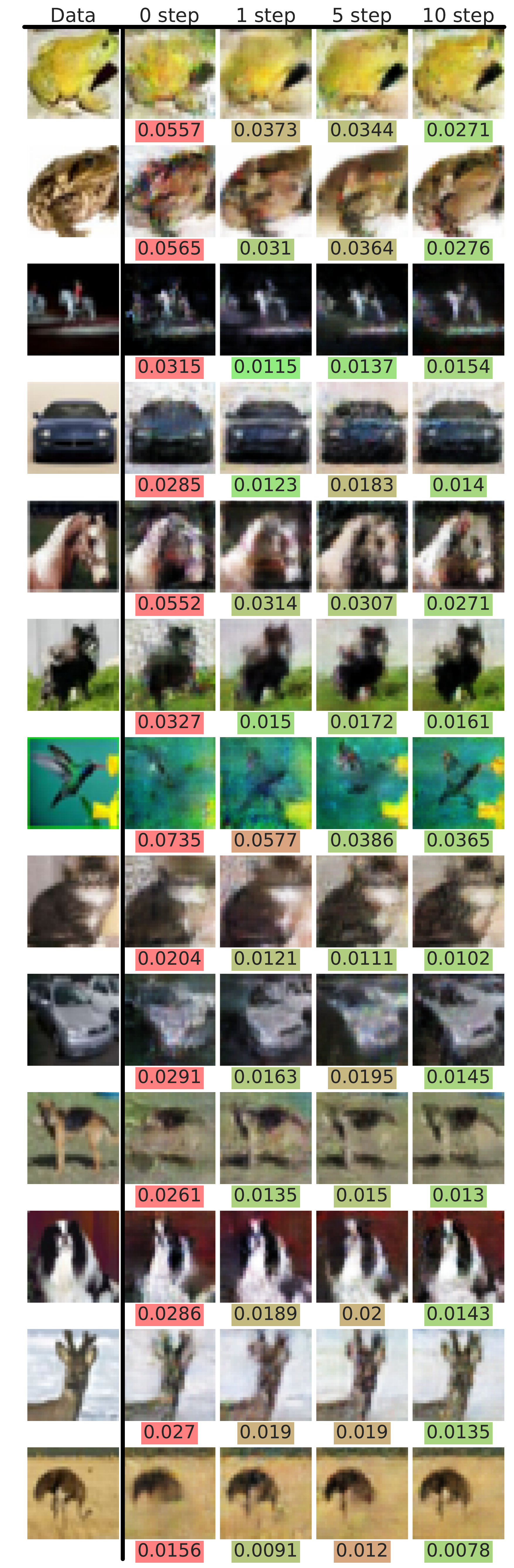}
  }
      \caption{Samples from 5/5 with different random seeds.}

\end{figure}

\end{document}